\documentclass[5p]{elsarticle} 
\usepackage[utf8]{inputenc}
\usepackage[T1]{fontenc}
\usepackage{url}
\usepackage{amsmath}
\usepackage{graphicx}
\usepackage{appendix}
\usepackage{todonotes}

\usepackage{float}
\title{Clustered Edge Intelligence: Beyond Just Convergence of Edge Computing and AI}
\author[one]{Chinmaya Kumar Dehury}
\ead{dehury@iiserbpr.ac.in}
\author[two]{Boris Sedlak} 
\ead{boris.sedlak@dsg.tuwien.ac.at}
\author[three]{Alaa Saleh} 
\ead{alaa.saleh@helsinki.fi}
\author[nine]{Ilir Murturi}
\ead{ilir.murturi@uni-pr.edu}
\author[six]{Lauri Lovén} 
\ead{lauri.loven@oulu.fi}
\author[five]{Satish Narayana Srirama}
\ead{satish.srirama@uohyd.ac.in}
\author[four]{Praveen Kumar Donta\corref{cor1}}
\ead{praveen@dsv.su.se}
\cortext[cor1]{Corresponding author}

\address[one]{Department of PComputer Science, IISER Berhampur, Odisha, India}
\address[two]{Distributed Systems Group, TU Wien, 1040 Vienna, Austria.}
\address[three]{Department of Computer Science, University of Helsinki, 00100 Helsinki, Finland}
\address[nine]{Department of Mechatronics, University of Prishtina, Prishtina 10000, Kosova.}
\address[six]{Center for Applied Computing, University of Oulu, 90100 Oulu, Finland}
\address[five]{School of Computer and Information Sciences, University of Hyderabad, India.}
\address[four]{Department of Computer and Systems Sciences, Stockholm University, 164 25 Stockholm, Sweden.}

\date{July 2026}
\journal{Information Fusion}
\begin{document}

\begin{abstract}
We are moving from an information age to the age of intelligence. A decade, or possibly less than that, data will not be the gold anymore rather the derived intelligence out of the data and the information we posses from the edge of the network. Existing Edge Intelligence research focuses mainly on two directions: using AI for edge resource management and deploying lightweight AI models on edge devices. However, existing edge-computing research lacks an intelligence-centric framework in which derived intelligence is treated as a first-class, independently manageable entity that can be described, discovered, observed, shared, reused, and dynamically clustered across heterogeneous edge devices and applications. To address these research gaps, we introduced Clustered Edge Intelligence (CEI), a visionary intelligence-centric approach. The aim of CEI is to make intelligence a shareable and reusable first-class entity that can be independently represented, discovered, observed, exchanged, and managed across the distributed edge–cloud continuum. We present a three-layer CEI architecture and examine enabling technologies and research dimensions, including intelligence inventories, semantic knowledge representation, communication, discoverability, observability, lifecycle automation, clustering mechanisms, marketplaces, interoperability, and standardization. 
\end{abstract}
\maketitle
\section{Introduction}
We are transitioning from the Information Age to the Age of Intelligence, where the saying ``data is the new gold'' will soon be a thing of the past. The spotlight will shift from merely managing and analyzing vast quantities of data---say zettabytes to exabytes to yottabytes---to managing and utilizing the derived intelligence. The process of transforming raw data into actionable intelligence involves a series of steps where data is contextualized into information, understood as knowledge, and ultimately applied to make informed decisions.

\paragraph*{Knowledge \& Intelligence} 
In the domain of computer science, knowledge refers to the information, facts, or understanding of the surrounding environment, including all tangible and intangible entities that we have acquired through a set of dedicated resource-constraint devices and networks \cite{rowley2007wisdom}. On the other hand, intelligence encompasses the cognitive abilities, reasoning, and learning capacity that make use of the existing knowledge \cite{russell2010artificial}.
Based on the Merriam-Webster\footnote{https://www.merriam-webster.com/dictionary/intelligence} dictionary, intelligence is \textit{the ability to learn or understand or to deal with new or trying situations}. 
In computer science, since decades a portion of the research revolving around embedding high-degree of intelligence onto the servers with high-end resource configuration. However, the research is slowly moving towards embedding certain level of intelligence to resource-constraints devices, e.g. the edge devices. 

\begin{figure*}[h]
    \centering
    \includegraphics[width=\linewidth]{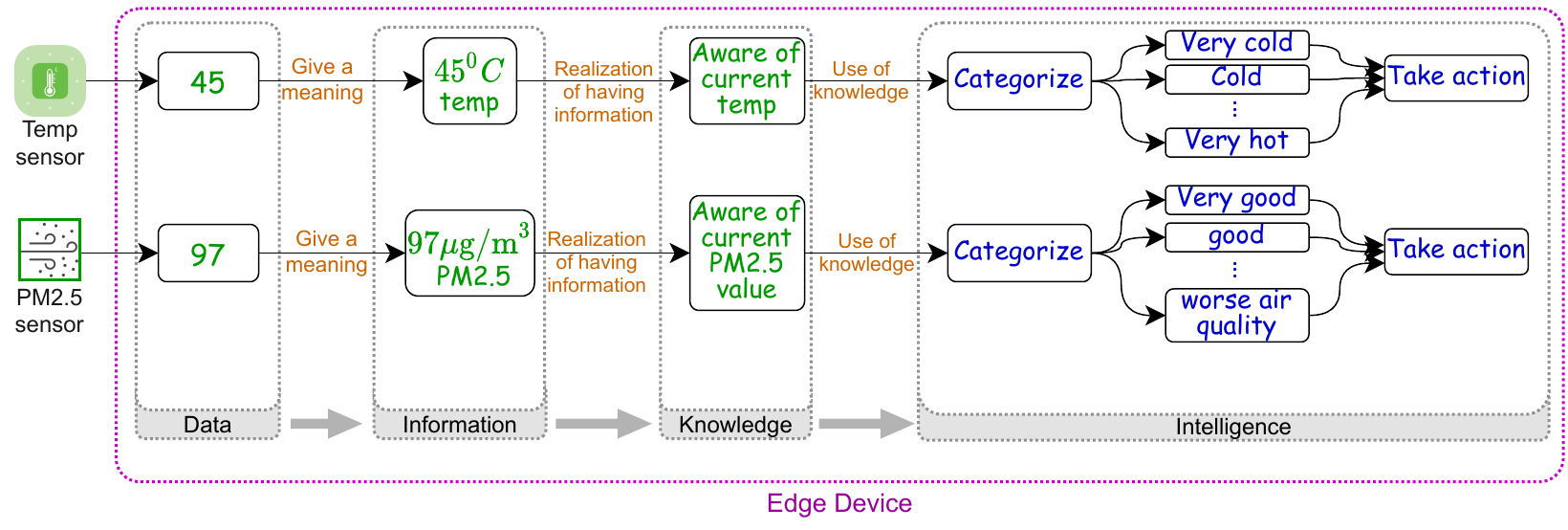}
    \caption{The transformation from Data to Intelligence: an example in Edge device}
    \label{fig:data-to-intelligence}
\end{figure*}

In the context of edge infrastructure, consisting of billions of edge devices, lets take an example and understand the transformation of data into intelligence, as shown in Figure \ref{fig:data-to-intelligence}. In this example, data is first collected from two sensors, say numerical value $45$ from a temperature sensor and $97$ from a PM2.5 sensor. Let's assume that both sensors are connected to one edge device. These raw data points, however, are devoid of context until they are given meaning, transforming them into information. For instance, the value $45$ is contextualized as $45^{\circ}$C temp, and $97$ is understood as $97 \mu\text{g}/\text{m}^3$ PM2.5, providing a clear understanding of what these numbers represent in the real world. This information is then processed further, where the device recognizes or becomes aware of the current environmental conditions, thus converting this information into knowledge. Knowledge in this context means that a system is not only aware of the current temperature and air quality levels but also understands their significance. Finally, this knowledge is used to categorize the environmental conditions into states such as \textit{Very Hot} or \textit{Good} air quality. 

Based on environmental conditions, the edge device may take appropriate action. For instance, in a smart home scenario, the edge device may directly communicate with the air purifier to turn on when the air quality is low. It may also directly instruct the air conditioning system to cool the environment. This final stage, where the system uses its knowledge to make decisions and act, represents the transformation of knowledge into intelligence. This demonstrates a clear distinction between knowledge and intelligence within the context of an edge infrastructure. Can we refer \textit{intelligence available at the edge of the network} as \textit{``Edge Intelligence''}? More onto this is discussed in Section \ref{sec:all_forms}.

\paragraph{Cohesion \& Coherence }
In general, the term \textit{Cohesion} can be defined as binding different elements/components to create a singular unit \cite{159342}. In the context of edge intelligence, Cohesion would refer to the integration and binding of intelligence available at different types of edge devices \cite{sheth2008semantic}. 
On the other hand, the term \textit{Coherence} can be defined as establishing a logical connection between diverse elements/components. In the context of edge intelligence, coherence can be seen as the connectedness/consistency that exists among intelligence possessed by different edge devices. 
For instance, a logical connection can be established between different sensors in the smart home scenario. The luminosity sensor and the proximity sensor are logically connected to the smart bulb. 
The level of CO2 is logically connected with traffic congestion and noise level. Higher noise levels on the road might indicate a higher value of CO2, and in a similar way, a higher value of CO2 may indicate higher noise pollution and more traffic congestion \cite{gately2017urban,espadaler2023traffic}. This shows a coherence or a logical relationship among different types of intelligence available at the edge of the network. In a very large-scale of edge infrastructure, where the number of edge devices may scale beyond few millions or billions,  management of those intelligence is a cumbersome task.

The current research is mostly progressing through clustering resource-constraint IoT devices \cite{8240666_2018_j2}, where clusters are formed mainly based on the geographic proximity, communication distance, sensing coverage, or proximity to an edge server \cite{achkouty2024rdsc,takele2025resource,begum2023data}. Such clusters are mainly responsible for performing a set of tasks with a single objective to achieve. This approach itself becoming a hurdle for implementing the edge intelligence that may involves a large number of IoT devices deployed in different geographical locations or of different types. For instance, with a conventional location-based clustering approach \cite{ghrab2023core}, it may not be possible to form a cluster of sensors that are available in the city buses. We may face the similar difficulties in case of clustering the sensors deployed in different solar panels in a smart city.

To address the above limitations, this paper introduces Clustered Edge Intelligence (CEI), an intelligence-centric vision where intelligence is considered as a first-class entity for its efficient  management across distributed computing continuum. The major contributions of this paper are summarized as follows.
\begin{itemize}
     \item To address the above limitations, this paper introduces Clustered Edge Intelligence (CEI), an intelligence-centric vision where intelliegnce is considered as a first-class entity for its efficient  management across distributed computing continuum. 
     \item We revisited the meaning of Edge Intelligence and classify it into three forms: AI for edge resource management, AI deployed at the edge, and intelligence available at the edge (our primary focus).
     \item CEI as a new paradigm is introduced where intelligence can be represented, discovered, observed, shared, reused, and clustered across heterogeneous edge devices. Following this, 
     \item we compared device-centric and intelligence-centric clustering and show how CEI decouples intelligence from the underlying hardware.
     \item A three-layer CEI architecture is proposed consisting of edge devices, edge controllers, and cloud environments.
     \item We identified key baseline technologies required for design and implementation of CEI, including abstraction, knowledge graphs, communication, discoverability, observability, clustering, and lifecycle-support mechanisms.
     \item Finally, we discussed major research dimensions and representative use cases to guide future research on intelligence-centric edge systems. 
\end{itemize}


\begin{figure*}
    \centering
    \includegraphics[width=0.95\linewidth]{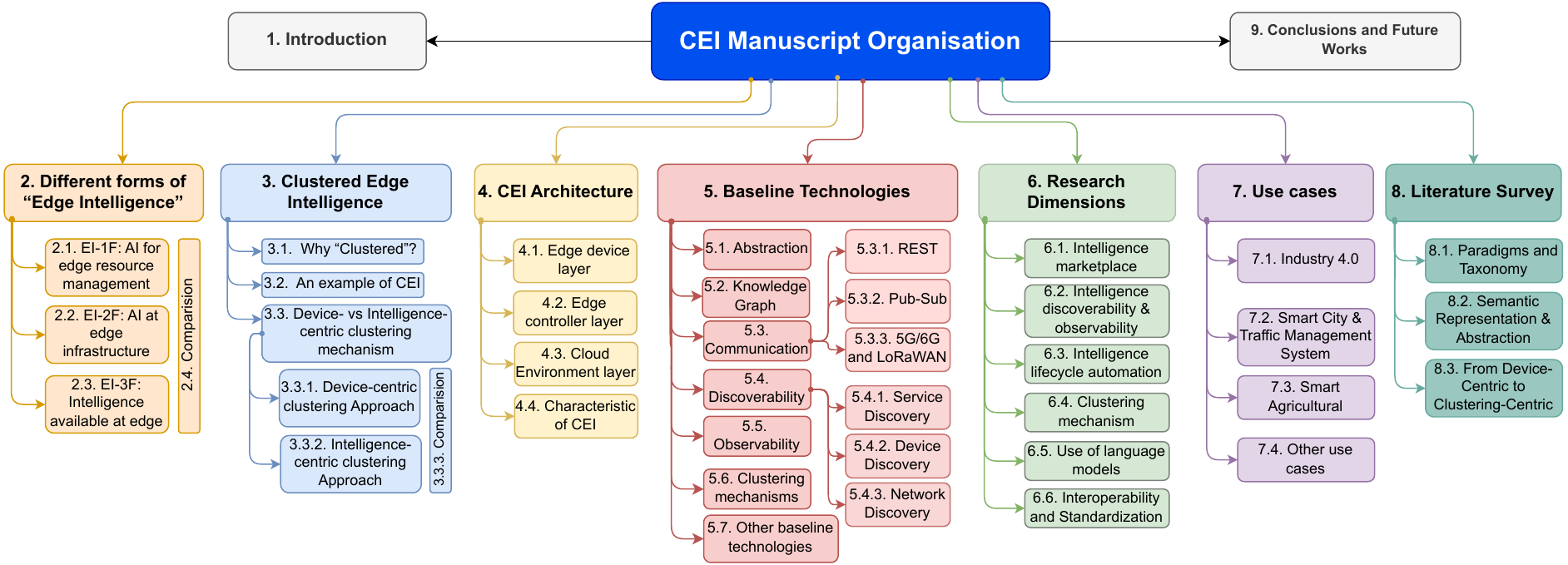}
    \caption{Organization of the manuscript}
    \label{fig:outline}
\end{figure*}

The remainder of this paper is organized as follows. In addition, for a quick reference, the overall outline of the manuscript is presented in Figure \ref{fig:outline}. Section \ref{sec:all_forms} discusses the different forms of Edge Intelligence, including AI for edge resource management, AI deployed at the edge, and intelligence available at the edge. Section \ref{sec:CEI_beyond_EC_AI} introduces the concept of Clustered Edge Intelligence, explains the motivation behind clustering intelligence, and compares device-centric and intelligence-centric clustering mechanisms. Section \ref{sec:CEI_Arch} presents the proposed CEI architecture, including the edge device layer, edge controller layer, cloud environment layer, and key system characteristics. Section \ref{sec:baseline} discusses the baseline technologies required to realize CEI, such as abstraction, knowledge graphs, communication mechanisms, discoverability, observability, clustering mechanisms, and other supporting technologies. Section \ref{sec:research_dir} highlights the major research dimensions, including intelligence marketplaces, intelligence lifecycle automation, clustering mechanisms, the use of language models, interoperability, and standardization. Section \ref{sec:usecases} presents representative use cases of CEI in Industry 4.0, smart city and traffic management, smart agriculture, and other application domains. Section \ref{sec:rel} provides the survey of existing research works. Finally, Section \ref{sec:conc} concludes the paper and outlines future research directions. Table \ref{tab:acronyms} provides the list of frequently used acronyms in this article. 

\begin{table}[!h]
    \centering\
    \footnotesize
    \caption{List of acronyms used within this paper.}
    \begin{tabular}{|p{0.2\linewidth}|p{0.7\linewidth}|}
        \hline
        \textbf{Acronym}	&  \textbf{Description} \\ \hline \hline 
        EI	& Edge Intelligence\\ \hline 
        EI-1F	& Edge Intelligence – First Form (AI for Edge Resource Management)\\ \hline 
        EI-2F	& Edge Intelligence – Second Form (AI at the Edge Infrastructure)\\ \hline 
        EI-3F	& Edge Intelligence – Third Form (Intelligence Available at the Edge)\\ \hline 
        ERM	& Edge Resource Management\\ \hline 
        EC	& Edge Computing\\ \hline 
        CEI	& Clustered Edge Intelligence\\ \hline 
        EA	& Edge Agent\\ \hline 
        DCC	& Device-Centric Clustering\\ \hline 
        ICC	& Intelligence-Centric Clustering\\ \hline 
        FL	& Federated Learning\\ \hline 
        RL	& Reinforcement Learning\\ \hline 
        IoT	& Internet of Things\\ \hline 
        API	& Application Programming Interface\\ \hline 
        REST	& Representational State Transfer\\ \hline 
        Pub-Sub	& Publish–Subscribe Communication Model\\ \hline 
        MQTT	& Message Queuing Telemetry Transport\\ \hline 
        HTTP	& Hypertext Transfer Protocol\\ \hline 
        IP	& intelligence provider\\ \hline 
        HAL	& Hardware Abstraction Layer\\ \hline 
        MPU	& Microprocessor Unit\\ \hline 
        EIM	& Edge Intelligence Marketplace\\ \hline 
        EIP	& Edge Intelligence Producer\\ \hline 
        ICC	& Intelligence Cluster\\ \hline 
        DD	& Device Discovery\\ \hline 
        D2D	& Device-to-Device\\ \hline 
        SD	& Service Discovery\\ \hline 
        SR	& Service Registry\\ \hline  
    \end{tabular}    
    \label{tab:acronyms}
\end{table}

\section{Different forms of ``Edge Intelligence''} \label{sec:all_forms}

\begin{figure*}[t]
    \centering
    \includegraphics[width=0.7\linewidth]{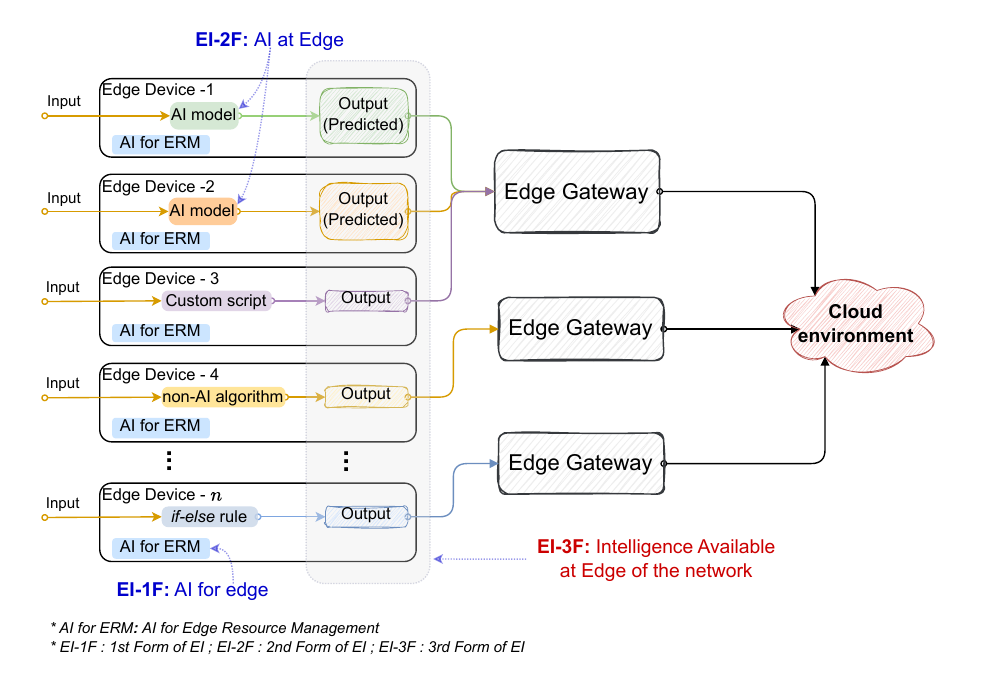}    
    \caption{Different forms of Edge Intelligence.}
    \label{fig:EI-meaning}
\end{figure*}

The term ``Edge Intelligence (EI)'' may have different meaning in different context and situations \cite{singh2023edge}. This can be referred to the design and deployment of AI algorithm for management of the edge computing, storage and network resources \cite{choudhury2024machine}. This definition is referred to as 1st Form (acronym EI-1F) in this article and used henceforth. On the other hand, in recent days the term ``Edge Intelligence (EI)'' is also refers to the minimalist version of an AI model (e.g. MobileNetV2~\cite{sandler2018mobilenetv2}, SqueezeNet~\cite{iandola2016squeezenet}, Edge Tensor Processing Unit
(EdgeTPU)-optimized models, YOLOv5-Nano~\footnote{\url{https://docs.ultralytics.com/models/yolov5}}, ShuffleNet \cite{zhang2018shufflenet}, etc) deployed at the edge of the network \cite{vyas_towards_2025}, as shown in Figure \ref{fig:EI-meaning}. In this article, this definition is referred to as 2nd Form (acronym EI-2F) and used henceforth. However, in this research, it is believed that the term EI may also refereed to the \textit{intelligence available at the edge} of the network, as shown in Figure \ref{fig:EI-meaning}, considered as 3rd Form (acronym EI-3F) and used henceforth. Such intelligence can be obtained using any AI, non-AI algorithms or any custom script. This can be realized in the context of smart city, supply chain, smart agriculture, and industry 4.0 real-world scenarios \cite{velaga2025edge}. Detailed explanation of all possible forms of EI and their comparisons are provided below.

\subsection{EI-1F: AI for edge resource management}
\label{sec:EI-1F}

Unlike the conventional approach of manual and brute-force approach of managing the edge resource management, the recent advancement of AI/ML is significantly integrated for the same purpose. AI is used in handling wide-range of tasks related to edge resource management, including network allocation \cite{10494796}, energy-aware scheduling \cite{electronics14204086}, task migration \cite{10480671}, compute allocation \cite{DEHURY2024111179}, upload scheduling and many more. 

For example, edge devices such as roadside CCTV analytics units or Jetson-based video processors often undergo thermal stress during continuous high-rate inference \cite{10666109}. A TinyML-based regression model can be embedded locally to predict short-term thermal load (e.g., 30–60 seconds ahead) based on current CPU/GPU utilization, ambient temperature, and inference rate. When the predicted temperature exceeds a critical threshold (e.g., 75°C), the system can autonomously downscale computation by lowering Frames Per Second (FPS), switching to a lighter AI model~\cite{sedlak_multi-dimensional_2025}, or reducing hardware clock speeds~\cite{10.1145/3801093}. Another example can be seen in network bandwidth allocation \cite{10494796}. Smart traffic poles frequently execute multiple competing tasks-such as license plate recognition, pedestrian detection, and environmental sensing-over bandwidth-limited 4G/5G uplinks. A lightweight reinforcement learning (RL) agent deployed within the pole can dynamically prioritize bandwidth allocation based on contextual demand \cite{9881565,9139661}. For example, under high traffic load, more bandwidth is allocated to CCTV analytics. 

Further, in remote isolated areas, solar-powered IoT nodes must preserve energy while performing the task assigned to it. A TinyLSTM (Tiny Long Short-Term Memory) model\footnote{\url{https://github.com/cpllab/tinylstm}} can predict future battery levels and solar input over the upcoming operational cycle. Based on these predictions, the node autonomously reduces sensor sampling frequency, disables non-critical sensing modules, or defers communication-intensive tasks when low-energy conditions are forecasted \cite{10.1007/978-3-031-82484-5_10}.


Through predictive analytics, AI can forecast edge and cloud resource failures, ensuring timely maintenance and reducing downtimes. Additionally, AI-driven security protocols can detect and mitigate potential threats, enhancing the robustness of the edge network. Furthermore, AI facilitates dynamic resource allocation, optimizing bandwidth and storage based on real-time data processing needs. In essence, the integration of AI into edge-cloud infrastructure management promises enhanced efficiency, security, and adaptability. Design of AI for edge device management is also referred to as the \textit{1st form of EI} in this article. 

\subsection{EI-2F: AI at edge infrastructure}\label{sec:EI-2F}
AI at the edge infrastructure, refers to the deployment of AI algorithms, especially the tiny version of AI, and minimalist models on the resource-constraint edge devices, close to the source of data, rather than in a centralized cloud-based system. This approach offers several advantages, including real-time processing, efficiency, speed, and enable autonomy. Lightweight AI models are deployed at the edge in a wide range of real-world applications. Albanese et al. \cite{albanese_automated_2021} highlighted the potential of edge AI in automating pest detection within fruit orchards, allowing for real-time monitoring and timely interventions. This enhances the overall efficiency of smart agriculture practices. Edge intelligence and deep learning are also applied in autonomous navigation in vineyards by Aghi et al. \cite{aghi_deep_2021}. The authors present a novel control mechanism that uses a custom-trained segmentation network and a low-range RGB-D camera to produce smooth trajectories and stable control in various vineyard scenarios. Radanliev et al. in \cite{radanliev_cyber_2020} discuss the integration of AI and edge computing in supply chains, emphasizing the mitigation of cyber risks, offering predictive cyber risk analytics with real-time intelligence from IoT networks at the edge. Recently Large Language Models (LLMs) are also being deployed on these edge devices providing intelligence for performing various natural language processing (NLP) tasks such as translation, question answering, text summarization, and generating clinical summaries from multimodal data such as text, sensor data, images and audio~\cite{srirama2026fogLLM}. This version of EI, also referred to as the \textit{2nd form of EI}, in this article.

\subsection{EI-3F: Intelligence available at edge }\label{sec:EI-3F} 
However, unlike the above-mentioned meanings of ``edge intelligence'', it can also be referred to as the ``\textit{intelligence available at the edge of the network}'', also known as \textit{ 3rd form of EI}. Some of the examples include intelligence related to energy production by Photovoltaic panels deployed atop city buses, intelligence related to trees, wildlife, and animals in the forest, and intelligence connected to soil conditions in the smart agriculture context. In the context of smart traffic, the camera not only captures edge knowledge (e.g. number of vehicles, their speed, traffic conditions, etc.) but also processes that knowledge on-the-spot to produce and share those intelligence (e.g. if a traffic is highly congested, if a city bus is delayed, if a vehicle is going beyond speed limit in a specific area, etc.) among other edge devices. Edge intelligence helps smarter decision making and can be used to enable predictive analytics and smarter services. In this article, our focus is to explore this version of EI, including the novelty, challenges, potential solutions, and baseline technologies or concepts. More on this is discussed on the later sections. 

\subsection{Comparison of three forms} 
Table \ref{table:EI_form_comparison} provides a comparative analysis of the three forms of Edge Intelligence (EI-1F, EI-2F, and EI-3F), taking some of the key parameters into account, such as hosting environment, purpose, challenges, development focus, deployment scenarios, complexity, interdependency, scalability, and security impact.

Unlike EI-1F and EI-2F, the primary goal of EI-3F (Intelligence available at Edge) is to obtain and share intelligence across devices to enable collaborative decision-making. However, achieving this goal introduces several challenges, including intelligence synchronization, distributed learning, intelligence discoverability \& observability, and maintaining consistency across devices.
The deployment scenarios of EI-3F are particularly suited for collaborative learning, where intelligence sharing among edge devices can mutually benefit multiple systems. A key example includes interconnected vehicles in a fleet, where each vehicle shares its intelligence to enhance navigation, safety, and traffic management in real-time~\cite{sedlak_slo-aware_2024_new}.

\begin{table*}[!t]
    \centering
    \footnotesize
\caption{Comparison of different forms of Edge Intelligence}
\label{table:EI_form_comparison}
    \begin{tabular}{|p{0.1\linewidth}|p{0.25\linewidth}|p{0.25\linewidth}|p{0.3\linewidth}|} \hline 
         \textbf{Parameter}& \textbf{EI-1F: AI for ERM}&  \textbf{EI-2F: AI@Edge}& \textbf{EI-3F: Intelligence available at Edge}\\ \hline 
         Hosting environment&  Edge Device, edge-cloud continuum, network devices &  Edge device& Edge Device, edge-cloud continuum, network devices\\ \hline 
         Purpose&  Edge computing and network resource management&  processing raw data available at edge& Obtaining \& sharing  intelligence, collborative decision-making \\ \hline 
         Challenges&  Making AI model tiny, limited computing resource, security, privacy \& Trust in AI-driven decisions & Limited computational resources, model optimization \& compression, lack of standardization, deployment complexity, hardware heterogeneity & Intelligence synchronization, distributed learning, intelligence discoverability, intelligence observability, consistency across devices \\ \hline 
         Development focus& - system optimization - edge infrastructure management - edge security  & model design, pruning, quantization, and other techniques to make AI models compact& - distributed learning - peer-to-peer intelligence exchange - collaborative learning.\\ \hline 
         Deployment scenario & This is required when (a)- dynamic edge resource allocation leads to performance gains (b)- dynamically securing edge device& where immediate action is required based on data - e.g. autonomous vehicles, health monitoring, or industrial automation.&- suitable for collaborative learning - when one edge device can be benifited from others - e.g. interconnected vehicles in a fleet\\ \hline 
         Complexity & Complex because of device and resource heterogenity & Complexity lies in making the model compact without compromising accuracy& complexity lies in - communication protocols - distributed system - heterogeneous network\\ \hline 
         Scalability & Not very challenging as the AI needs to manage single edge device's resources& not very challenging, as the AI models only concern about data generated by that device only & very challenging\\ \hline 
         Security impact &  The entire edge infrastructure&  users' applications & users' applications\\ \hline
    \end{tabular} 
\end{table*}

\section{Clustered Edge Intelligence: Beyond EC and AI convergence}\label{sec:CEI_beyond_EC_AI}

Edge Intelligence (EI) in general refers to the convergence of edge computing and artificial intelligence \cite{qiu_bring_2020, rausch_edge_2019, plastiras_edge_2018, peltonen_6g_2020, molokomme_edge_2022, parekh_edge_2021_new}, be it \textit{AI for edge} or \textit{AI on edge}. In conventional EI, the AI/ML algorithms are deployed on edge computing node, which complement its counterpart cloud deployment to provide realtime decision-making and predictive analytics service, as also discussed in Section \ref{sec:EI-1F} and \ref{sec:EI-2F}. For example in a modern traffic control system, the identification and blurring of a person's face can be done at the edge using light-weight AI algorithms~\cite{sedlak_privacy_2023}. One of the major advantages of EI is to perform AI/ML tasks and data analysis locally on the device, instead of sending all data to a central server, which can reduce latency, maintain privacy, and reduce network bandwidth consumption. Some of the applications of conventional EI includes: 
\begin{itemize}
    \item Smart Home: The devices such as smart speakers, thermostats, and security cameras can use EI to analyze smart home sensor data and make decisions locally, such as adjusting temperature and indoor luminosity, notifying wastage of electricity or alerting homeowner the potential security threats, etc \cite{thakur2024edge}.
    \item Industry 4.0: EI can be used in modern manufacturing industries to analyze robot and machinery sensor data to detect anomalies, predict failure and maintenance needs, optimize production pipeline, etc \cite{savaglio2024edge}.
    \item Smart Traffic and Vehicular network: With EI a wide range of traffic management tasks can be performed, including CCTV footage analysis, predict traffic flow, congestion, and accidents \cite{shen2024traffic}. With EI vehicles can exchange information about road conditions and traffic. In autonomous driving, EI enable sensors data analysis in real-time manner, make decisions about steering, braking, and acceleration. This can play an important role in improving the safety and reliability of autonomous driving \cite{qi2021extensive}. 
    \item Retail: EI allows the retailer to analyze data from sensors and cameras to implement smart shelves, track customer behavior, optimize checkout process, detect fraud, and improve inventory management.  
\end{itemize}

Under the current approach, edge devices remain highly dependent on a central system, with limited or negligible opportunities to share their learning directly with peer edge devices. For instance, in Federated Learning (FL), multiple edge devices contribute to improving and updating a central machine-learning model while keeping their training data locally \cite{mcmahan2017communication, li2025federated}. This approach helps achieve several important objectives, including reduced latency, improved reliability and scalability, enhanced privacy and security, lower network bandwidth consumption, reduced service cost, and improved service quality. However, it does not fully address the research question of how edge devices with similar characteristics, capabilities, or configurations can directly share their learning or derived intelligence without relying on a central system. Conventional Edge Intelligence research therefore provides limited support for peer-level collaboration, transfer of learning and knowledge, and collaborative reasoning. To address this gap, we revisit the concept of Edge Intelligence from a different perspective and interpret it as ``Intelligence Available at the Edge'', the 3rd form of EI (EI-3F), as briefed in Section \ref{sec:EI-3F}.

In EI-3F, intelligence is treated as a first-class entity that can be uniquely identified, discovered, observed, shared, and reused. Edge agents (EAs) deployed on edge devices are responsible for deriving intelligence from locally available data and knowledge. An edge agent is an autonomous software entity operating on an edge device that processes local data and knowledge, derives and publishes intelligence, discovers and consumes external intelligence, and collaborates with other edge agents to support context-aware decision-making.

An edge agent consists of three primary components: an AI model or algorithm, metadata, and business logic. The metadata may describe the type of data processed, location, data-generation frequency, communication requirements, and other contextual properties. The business logic defines how the data is used, how the agent communicates with other edge agents, and how contextual awareness is incorporated into decision-making. In conventional approaches, these components are often hard-coded and tightly coupled with the underlying host device. In the proposed approach, the edge agent encapsulates these components and decouples them from the host device, thereby improving portability, reusability, and manageability. Edge agents are also responsible for producing intelligence and managing its lifecycle, including publication, update, monitoring, and retirement. Multiple edge agents may be co-located on the same edge device; however, a single edge-agent instance is assumed to operate on one edge device at a given time~\cite{sedlak_multi-dimensional_2025}. This raises an important question: if intelligence is treated as a first-class entity, why is there still a need to cluster multiple intelligences?


\subsection{Why ``Clustered''? }

Traditionally, intelligence is derived based on specific information, mostly in the cloud computing environment. For example, whether a road is congested or not is typically determined in a cloud environment after receiving location information from multiple vehicles or analyzing video footage. These are classic examples where intelligence is derived from the same type of data. However, we believe that multiple types of information can be used to derive more \emph{reliable} intelligence. For example, location data from vehicles, CCTV footage, and $CO_2$ levels in a specific road segment can be combined to more accurately and reliably determine traffic congestion. Extending this to a more practical scenario, we propose that multiple types of intelligence, produced by edge agents deployed on edge devices, can be semantically clustered together to generate more complex intelligence. The following subsection provides a more concrete example to simplify the concept behind Clustered Edge Intelligence (CEI).




\subsection{Clustering Edge Intelligence : An Example }
\begin{figure*}[t]
    \centering
    \includegraphics[width=0.75\textwidth]{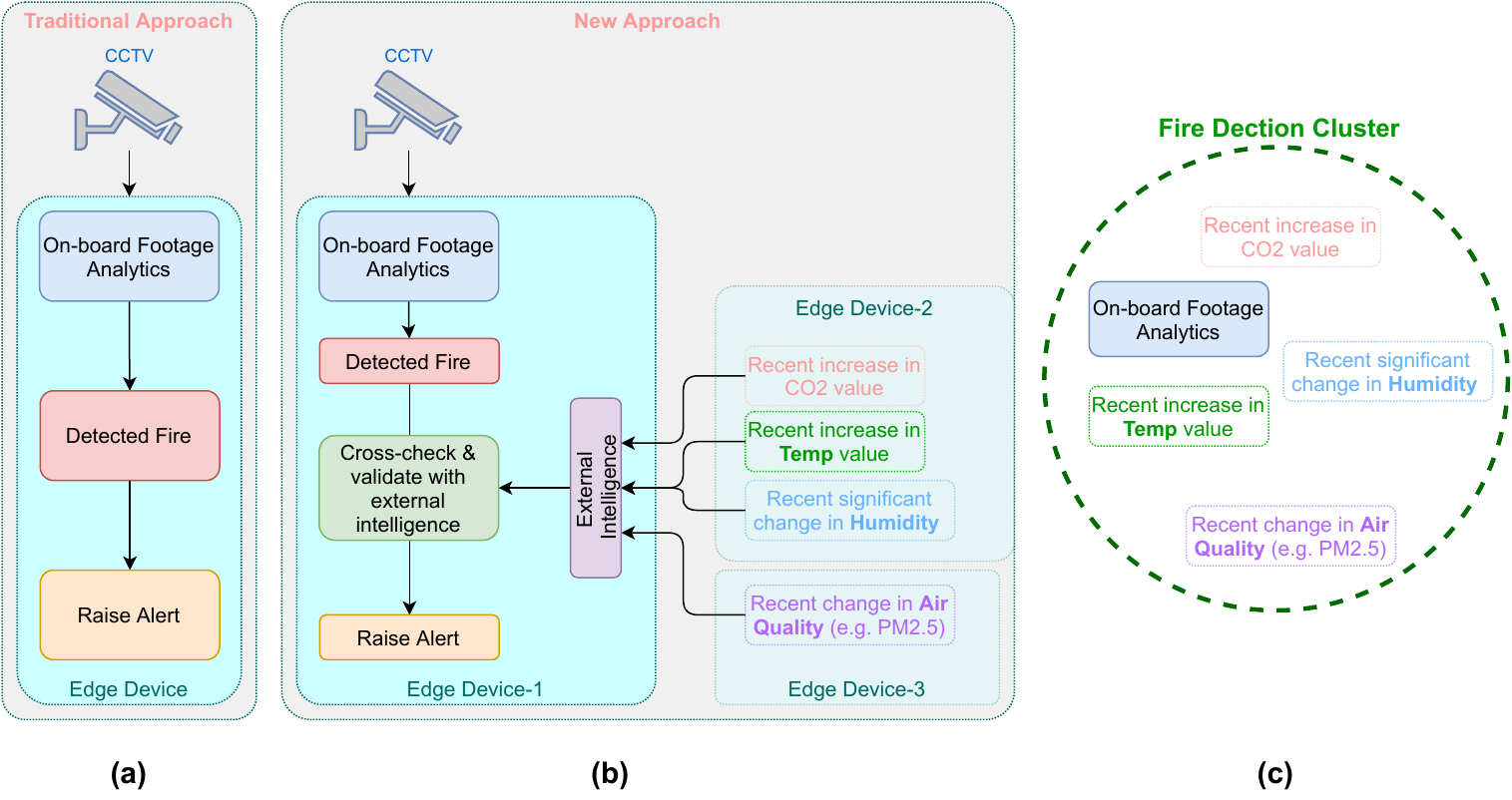}
    \caption{Comparison between traditional and CEI-enabled fire-detection mechanisms (an example): (a) single-source detection, (b) intelligence-assisted validation, and (c) logical clustering of related intelligence for reliable fire detection.}
    \label{fig:CEI-Example}
\end{figure*}

Let's explore the benefits of Clustered Edge Intelligence (CEI) with an example of a remote fire detection and alert system, as shown in Figure \ref{fig:CEI-Example}. In the conventional approach, as in Figure \ref{fig:CEI-Example}(a), the setup relies solely on one type of edge sensor, combined with onboard data analytics, say CCTV camera sensor combined with on board footage analytics. Onboard footage analytics (which could be a tiny AI model or specialized computer-vision algorithm) continuously captures the video footage from CCTV camera to detect visual indicators of fire within the view-point of the camera. If a fire is detected, the system immediately triggers an alert, notifying the relevant authorities or systems to respond. This approach is straightforward but has limitations, primarily because it depends entirely on the visual detection of fire. As a result, it may be prone to false alarms or could miss fires that do not have clear visual signs, such as those covered by obstacles or occurring in areas not covered by the camera.

However, with the CEI approach (Figure \ref{fig:CEI-Example}(b)), the fire detection mechanism can be further enhanced by integrating additional layers of intelligence and data validation before raising an alert. Instead of raising an alert immediately upon detecting fire, the edge device can cross-check and validate the detection with external intelligence. This external intelligence may come from other edge devices. For instance,  the EA deployed on \textit{Edge Device - 2} may provide environmental intelligence such as recent change in CO2 levels, temperature, and humidity. The EA on \textit{Edge Device - 3} may provide intelligence if there is any recent significant change in Air Quality. By comparing the footage-based fire detection with real-time external environmental intelligence, the agent on \textit{Edge Device -1} can more accurately and reliably confirm whether a fire is actually occurring. Only after this validation process is complete the system raise an alert. This approach significantly reduces the likelihood of false alarms and ensures a more reliable response to potential fire hazards.

Through the CEI (Clustered Edge Intelligence) approach, we can establish a cluster of related intelligence produced by various agents. As illustrated in Figure \ref{fig:CEI-Example}(c), this cluster integrates intelligence such as recent changes in CO2 levels, temperature, and humidity from \textit{Edge Device-2}, along with significant recent changes in air quality from \textit{Edge Device-3}. It's important to note that these edge devices do not share raw data; instead, the deployed EAs share derived intelligence based on the data available to each respective device. This shared intelligence is collectively analyzed to assess the likelihood of a fire, enhancing the system's ability to detect and respond to potential threats effectively.

At this point, one might wonder how intelligence clustering is different or similar to the clustering of underlying edge devices. This also raises another question: why to form clusters based on the relativity of intelligence rather than the devices that provide it? We will address both questions in detail in the following subsection.

\subsection{Device- vs Intelligence-centric clustering mechanism}
In continuation to explore the concept of clustered edge intelligence and our previous work on clustered and cohesive edge intelligence \cite{dehury2022ccei}, in this subsection, we compare two different clustering approaches: Device-centric clustering and Intelligence-centric clustering approaches. Figure \ref{fig:dcc-vs-icc} presents a comparative analysis of two different clustering approaches.

\subsubsection{Device-centric clustering (DCC) Approach:}
\begin{figure*}
    \centering
    \includegraphics[width=0.85\textwidth]{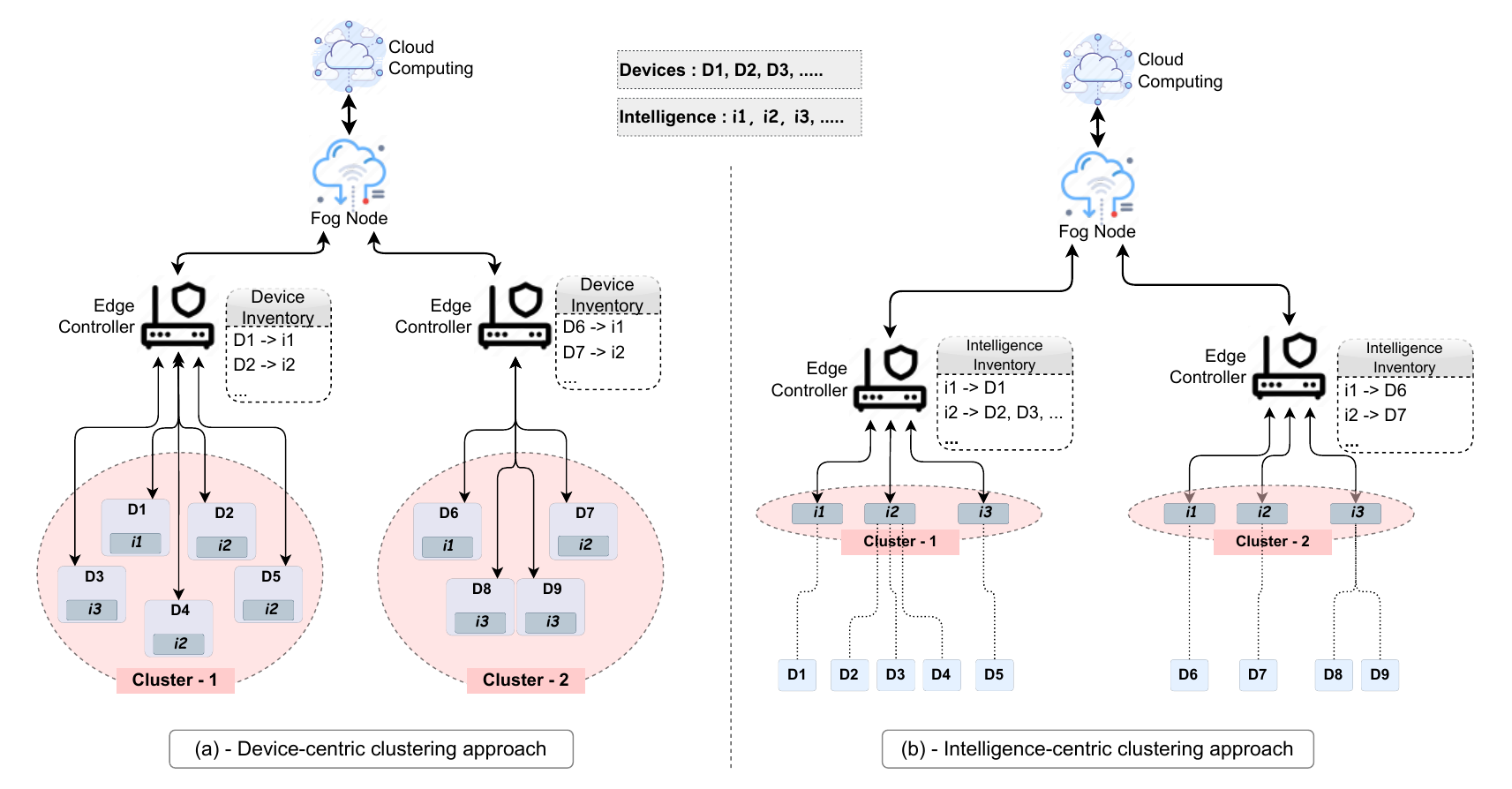}
    \caption{A pictorial representation of (a) Device-centric clustering approach, and (b) intelligence-centric clustering approach.}
    \label{fig:dcc-vs-icc}
\end{figure*}

In device-centric clustering approach (Figure \ref{fig:dcc-vs-icc}(a)), clusters of physical edge devices are formed based on their physical characteristics, proximity, or functional roles, rather than the specific intelligence they generate. Each cluster contains a group of devices that are managed by an edge controller, which communicates with a fog node connected to cloud computing resources. The edge controller maintains a device inventory, mapping each device (e.g., $D1$, $D2$) to the specific type of intelligence (e.g., $i1$, $i2$) it generates. For instance, devices $D1$ and $D2$ generate intelligences $i1$ and $i2$, respectively. This approach organizes the devices into clusters without considering the specific type of intelligence they produce, instead of primarily focusing on the management of the devices themselves.

Each device within the cluster is usually identified by a unique identifier. The intelligence could either be stored directly in the inventory or queried from the device in real-time when needed. Accessing the intelligence in a device-centric clustering approach requires the user to have knowledge of specific device identifiers. This means that before any intelligence (such as ``What is the percentage of changes in $CO_2$ value in last $10 mins$ in location X?'') can be accessed, the user must first ascertain whether the desired device is present within the network. The process begins with the user sending a query to the edge gateway, which serves as the management hub for the device cluster. The query typically asks whether a particular device is available in the inventory. If the device is present, the edge gateway retrieves the surrounding intelligence from that specific device. Further, in a traditional approach, the intelligence is computed/derived on-demand either on the source edge device or computed on the edge device where the request is generated. 

An important aspect of this approach is the tight coupling between devices and the intelligence they generate. This tight coupling means that the device-centric approach emphasizes the management and availability of physical devices. To access a particular piece of intelligence, it is crucial first to ensure that the device capable of providing that intelligence is available in the network. This process is repeated for all devices connected to the edge gateway, making the device the primary focus of the clustering and intelligence retrieval processes.

It reflects a more hardware-oriented view of edge computing, where the infrastructure and devices themselves are central, and the intelligence they provide is secondary. This approach can be advantageous in scenarios where specific devices are known to generate unique and critical intelligence.  The primary challenge with this approach is the risk of vendor lock-in. Different vendors may use distinct data models to provide data or intelligence, leading to compatibility issues. Furthermore, if multiple devices offer the same type of intelligence, it becomes difficult to seamlessly substitute one device with another from a different vendor in case of failure or unavailability. This lack of interoperability can restrict the effective use of alternative devices, limiting flexibility and resilience in the system.

\subsubsection{Intelligence-centric clustering (ICC) Approach:}

On the contrary, as proposed in this research work, in the intelligence-centric clustering approach (Figure \ref{fig:dcc-vs-icc}(b)), the fundamental difference is that the focus is not primarily on device management but on the management of  intelligence available. There is a shift from focusing on the physical devices to prioritizing the intelligence they generate. In this approach, the edge controller maintains an intelligence inventory, mapping each type of intelligence (e.g., $i1$, $i2$) to the corresponding devices that provide it. For example, intelligence $i1$ might be derived from device $D1$, while intelligence $i2$ could be generated by one or more number of devices from multiple devices such as $D2$, $D3$, and $D4$. In this approach, clusters (Cluster-1 and Cluster-2) are formed by grouping together intelligences rather than devices. It is to be noted that, the inventory may include intelligence for which there are currently no connected devices, highlighting the system's flexibility and its decoupling of intelligence from specific hardware.

When a user seeks to obtain certain environmental intelligence (e.g., last $30 min$ average humidity on location A, Air quality ($PM2.5$ value) improved by X\% in last $24 hours$, etc.), they send a query to the edge controller. The controller response process is streamlined: it checks its intelligence inventory to determine if the requested intelligence is within its scope. If the intelligence is available, the controller queries the appropriate device and forwards the resulting data back to other edge devices or even the end-user. It is not required to have any knowledge of the specific devices providing the intelligence. The management of devices is kept abstract, meaning, details such as the device’s age, vendor, operating frequency, or other hardware-related characteristics are abstracted out from other edge devices or the end-user. This abstraction simplifies the user experience, allowing them to focus solely on the intelligence they need rather than the complications of the devices that produce it.

Another important aspect of this approach is the flexibility it offers. Users querying the same edge controller for the same intelligence at different times might receive data from different devices, depending on which devices are available and capable of providing the intelligence at that moment. This ability to source the same intelligence from multiple devices reduces dependency on any single device and enhances the system's resilience. If a user receives no response to their intelligence query, it could indicate that the relevant device is unavailable, or it might be due to security restrictions that prevent access to the intelligence.

The intelligence-centric approach is characterized by a loose coupling between devices and the intelligence they generate. This decoupling allows for greater flexibility and scalability, as intelligence can be accessed from any suitable device without the user needing to manage or even be aware of the underlying hardware. 

\subsubsection{Comparison of device-centric and intelligence-centric clustering}
Table \ref{tab:dcc-vs-icc} provides a comparison of Device-Centric Clustering (DCC) and Intelligence-Centric Clustering (ICC) approaches, taking several aspects into account, as summarized below. 
\begin{table}[h]
\centering
\footnotesize
\caption{Comparison of Device-centric clustering (DCC) \& intelligence-centric clustering (ICC) approach}
\begin{tabular}{ |p{0.25\linewidth} |p{0.35\linewidth}|p{0.27\linewidth}| }
\hline 
  \textbf{Parameters} & \textbf{DCC} &  \textbf{ICC} \\ \hline
 Device Hardware and Intelligence Coupling& very tightly coupled &very loosely coupled \\ \hline 
 Vendor lock-in issue & hard to avoid & can be easily avoided \\ \hline 
 Management & Focus on management of devices. Intelligence wil not be affected.& Focus on management of intelligence. Devices wil not be affected.\\ \hline 
 Clustering strategy implementation & implemented at device level & implemented at Edge gateway level \\ \hline
 Discoverability \& observability& focus is on edge devices& Focus is on edge intelligence\\ \hline
 Security \& privacy& Does not help in enhancing security and privacy&  Helps in enhancing security and privacy\\ \hline
 \end{tabular}\label{tab:dcc-vs-icc}
\end{table}
\begin{itemize}
    \item \textit{Device Hardware and Intelligence Coupling}: DCC shows a ``very tightly coupled'' relationship between device hardware and the intelligence derived, meaning that the intelligence is often inseparable from the device that produces it \cite{silva2024chameliot}. On the other hand, in ICC,  intelligence is abstracted from the hardware, making it very loosely coupled.
    
    \item \textit{Vendor Lock-in Issue}: In DCC, the vendor lock-in issue is ``hard to avoid'' due to the tight coupling between devices and the intelligence they generate. Each vendor might use proprietary data models and standards, making it difficult to integrate devices from different vendors \cite{chan2023iot,hoglund2024autopki}. In contrast, in ICC this issue can be easily mitigated because intelligence is loosely coupled with devices, which makes it interoperable across different vendor devices.
    
    \item \textit{Management of Device and Intelligence}: The management of devices in DCC directly affects the associated intelligence, as the two are tightly linked. Conversely, in ICC, while intelligence management is crucial, it does not directly impact the management of devices. This makes the system management more independent and flexible.  
    
    \item \textit{Clustering Strategy Implementation}: In DCC, clustering is implemented at the device level, requiring updates and coordination across all devices involved in the cluster. In ICC, the clustering strategy is independent of the device and implemented at the Edge gateway or controller level.  
    
    \item \textit{Discoverability and observability}: In DCC, during the cluster formation process, the discoverability and observability mechanisms primarily focus on the devices themselves, regardless of the specific intelligence they are capable of. In contrast, ICC shifts the emphasis in discoverability and observability to the intelligence, rather than the underlying devices. This approach in ICC prioritizes the accessibility and management of intelligence, treating the devices as secondary to the information they produce.
    
    \item \textit{Security and privacy}: In DCC, security and privacy rely on the imposed security mechanisms, which may involve sending sensitive data to the cloud \cite{pathak2024securing}. On the other hand, in ICC, the security and privacy can be enhanced in a way that the intelligences are derived on-board and only shared with others. The system is not promoting the sharing of data with other edge devices. 
\end{itemize}

\section{CEI Architecture}
\label{sec:CEI_Arch}

\begin{figure}[h]
    \centering
    \includegraphics[width=0.80\linewidth]{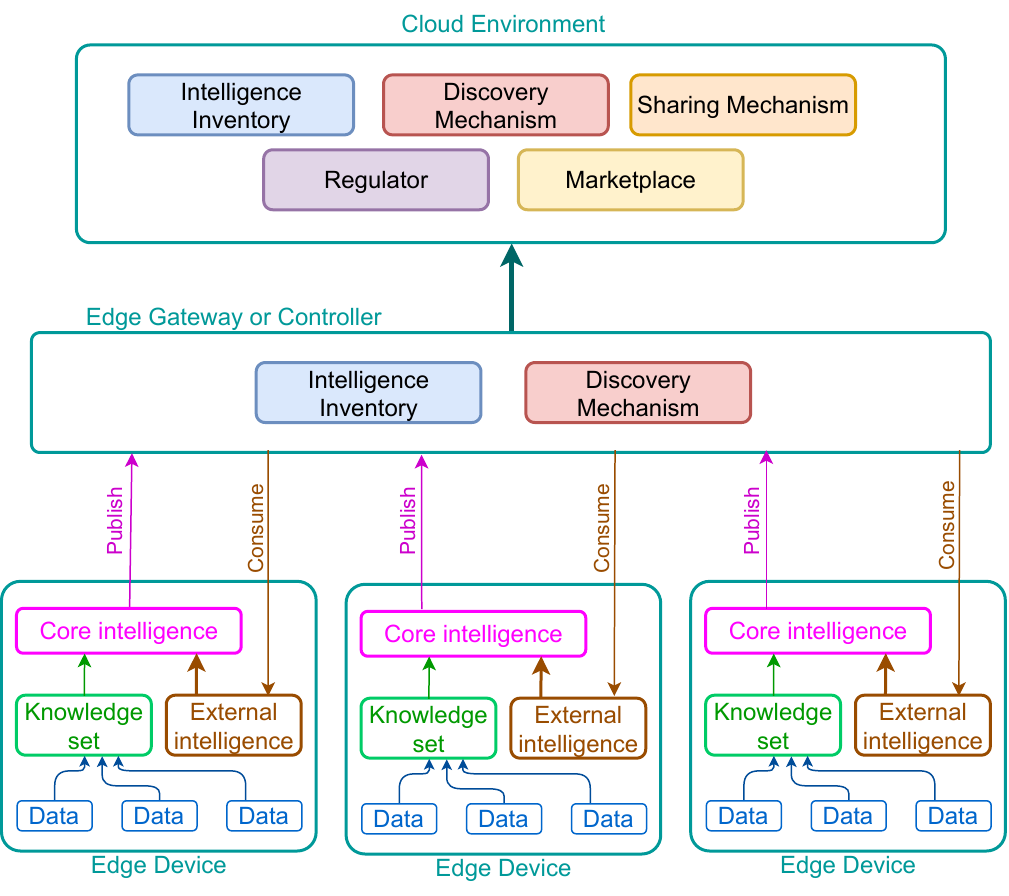}
    \caption{Overall Architecture of CEI}
    \label{fig:CEI_arch}
\end{figure}

Figure \ref{fig:CEI_arch} illustrates the three-tiered architecture of Clustered Edge Intelligence (CEI), consisting of edge devices at the foundational layer, an edge gateway or controller in the intermediary layer, and the cloud at the top. The architecture’s core objective is the effective management of intelligence within the edge infrastructure. Edge devices exhibit two forms of intelligence: \textit{Core intelligence}, derived from their own data, and \textit{External Intelligence}, which integrates insights from other devices.

\subsection{Edge device layer}
\begin{figure}
    \centering
    \includegraphics[width=0.9\linewidth]{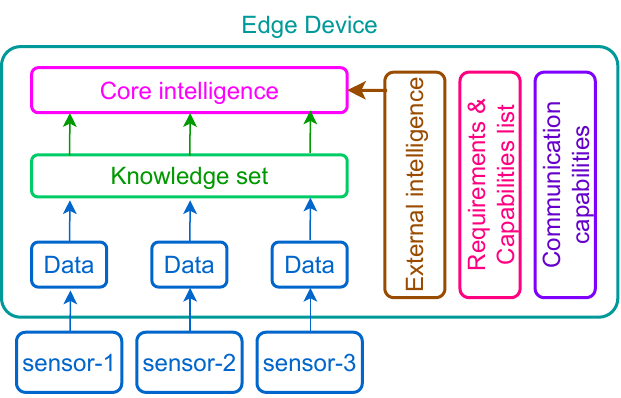}
    \caption{A detailed view of edge device in CEI.}
    \label{fig:CEI_edge_device}
\end{figure}

Figure \ref{fig:CEI_edge_device} showcases the internal components of an individual edge device within the CEI architecture. Each edge device is connected with multiple sensors, collecting environmental data such as temperature, humidity, noise levels, and luminosity at varying intervals. These sensors are either embedded on-board or wirelessly connected via technologies like Long Range Wide Area Network (LoRaWAN), Wi-Fi, or Bluetooth. The edge device receives the data,  processes, interprets, and contextualizes each data point (e.g., identifying whether the value ``40'' refers to temperature, humidity, or another metric). It is possible that an edge device can be connected to multiple sensors. In such a case, the edge device will be equipped with different varieties of knowledge sets.

Further, as discussed before, limited computing resources are available on each edge device. With such limited computing resource, edge devices are capable of performing several tasks, including processing specific data types, acting on data, executing cloud-based instructions, communicating with other devices, and discovering peers. Conversely, edge devices have ``Requirements,'' such as accessing core intelligence from other devices, retrieving intelligence from shared Intelligence Inventories (discussed later), or acquiring specialized knowledge from peers. Additionally, edge devices are aware of their \textit{Communication Capabilities}, including Wi-Fi, Bluetooth, or LoRaWAN, enabling connections with multiple peers across different channels.

Each edge device is designed to share its intelligence with others while also utilizing intelligence from other devices. When an edge device receives information from its peers, it is called \textit{external intelligence}. For example, an edge device connected to a CCTV camera can capture footage of its surroundings and detect signs of a fire. At the same time, it may receive external intelligence from nearby devices, such as a sudden increase in temperature. By cross-verifying this information, the device can confirm the presence of a fire with greater confidence. This verified conclusion can also be referred to as \textit{core intelligence}.

\subsection{Edge controller layer}
As mentioned earlier, each edge device is designed to share and collaborate with others. However, the key question is: How can they effectively communicate, collaborate, and share their knowledge and intelligence? The answer lies in Figure 5, which illustrates the overall architecture of Clustered Edge Intelligence (CEI). To enable communication between edge devices, the cloud, and their peers, an edge gateway (or controller) is introduced as an intermediary layer. The edge controller is equipped with two key components: the Intelligence Inventory and the Discovery Mechanism.

The \textit{Intelligence Inventory} is responsible for maintaining a record of all available intelligence across connected edge devices. This intelligence is typically shared by multiple devices. For example, an intelligence record may indicate that traffic is 50\% congested or that a road is blocked due to an accident. Each record includes essential metadata shared by the edge devices, such as:
\begin{itemize}
    \item The intelligence (e.g. \emph{``amount of electricity produced by solar panels in all city buses''}, \emph{``level of CO2 at any given point of time''}, \emph{``level of noise in a specific city area''}, etc.)
    \item Category (e.g., temperature, humidity, $CO_2$, PM2.5, soil condition, traffic condition, road condition, etc.)
     \item Access URL (to retrieve intelligence)
     \item Input and output format (which defines how another edge device should structure its message and what format to expect in response)
     \item Confidence score (indicating the reliability of intelligence)
     \item Application protocol (e.g., MQTT, CoAP, REST API)
     \item Communication channel (e.g., LoRaWAN, Bluetooth Low Energy (BLE), Wi-Fi)
\end{itemize}

The \textit{Discovery Mechanism} is designed to allow edge devices to access external intelligence shared by others. Unlike traditional IoT/edge infrastructure, which focuses on discovering physical devices, this mechanism is dedicated to discovering intelligence. If an edge device requires intelligence that is not available locally, the discovery mechanism searches for relevant intelligence across nearby edge devices or intelligence clusters. This enables peer-to-peer intelligence sharing without relying on cloud intervention. It is important to note that the \textit{Core Intelligence} component within the edge device is responsible for publishing intelligence along with its essential metadata. Meanwhile, the \textit{External Intelligence} component is responsible for consuming shared intelligence by communicating with the edge controller.

\subsection{Cloud Environment layer}
An edge controller has limitations in communicating with and controlling a large number of edge devices. In a large-scale edge infrastructure, there could be thousands or even millions of edge controllers. The cloud environment at the top, is equipped with a global view and is responsible for managing and coordinating all edge controllers. Similar to an edge controller, the cloud layer also consists of an  Intelligence Inventory and a Discovery Mechanism components. The discovery mechanism enables an edge controller to identify and access intelligence from other edge controllers' intelligence inventories. Additionally, the cloud layer serves as a global, centralized intelligence repository.

Furthermore, with the \textit{Intelligence Marketplace} component, the cloud functions as a marketplace where organizations-whether they own AI models, edge devices, or data-can buy, sell, and exchange AI models, decision rules, and other intelligence. The \textit{Regulator} component within the cloud enforces access control policies, security restrictions, and compliance rules to prevent unauthorized intelligence sharing. The \textit{Sharing Mechanism} provides necessary instructions and policies to edge devices, defining what intelligence should be shared, with which peers, and how frequently.


\subsection{CEI Characteristics}
The effectiveness of CEI depends on a set of core characteristics that make the edge agents independent, adaptable/dynamic, capable of exchanging contextual information, and collaborate with other agents. Some of the primary characteristics are summarized below:
\begin{itemize}
    \item \textit{Dynamicity of the environment}:  In such a large-scale distributed edge systems, devices and agents may frequently join or leave the network due to mobility (e.g., drones, autonomous vehicles), power constraints, network availability, or hardware failures. This dynamic nature demands that edge agents be context-aware and capable of adjusting their behavior in real time - for instance, re-routing communication paths when a node goes offline, or switching from visual to audio input when visibility is poor.
    \item \textit{Autonomy}: The fundamental and inherent characteristic of edge agents in such a distributed system is the autonomy, which refers to the ability of each edge agent to operate independently, understand the data and derive all possible intelligence related to that data/information, and execute tasks without requiring constant supervision from a cloud-based central entity. The autonomy also in the context of discovering other edge agents with different capabilities. For example, in a smart agriculture scenario, an edge agent responsible for crop identification may need to discover and coordinate with another edge agent-possibly on a different drone-tasked with pesticide spraying, without relying on central oversight.  
    \item \textit{Context Awareness \& Sharing}: Context awareness allows each agent to interpret its environment, understand its current state, and recognize the relevance of that information to a shared task. For effective collaborative learning, this becomes  important when multiple edge agents are capable of performing the same or similar tasks, but in different environmental or situational contexts. For example, in a Smart Agriculture setting, edge devices such as soil moisture sensors, drones, and weather stations operate across different fields and zones. A drone monitoring crop health in one area might detect signs of pest infestation. By sharing this context with nearby drones or ground sensors, those agents can proactively adjust their scanning or sampling behaviors, even if they operate in slightly different microclimates or crop types. This shared situational awareness enhances the overall system’s intelligence and its reliability.
    \item  \textit{Inter-Agent Communication Protocols}: Because edge environments are resource-constrained and often operate under dynamic network conditions, communication must be lightweight, resilient, and low-latency \cite{DIB2026104577}. Common communication paradigms include publish-subscribe models (e.g., MQTT), remote procedure calls (e.g., gRPC), and peer-to-peer messaging using custom APIs or agent frameworks. Edge agents should be capable enough to take advantage of available communication channel to announce their presence, share capabilities, request assistance, or synchronize on tasks. For example, a fall detection AI on a wearable device might send an alert to a nearby surveillance camera to verify the event visually before escalating. 
\end{itemize}


\section{Baseline Technologies}\label{sec:baseline}
The implementation of the CEI architecture and its capabilities requires a set of enabling technologies that support the representation, communication, discovery, management, and clustering of intelligence across distributed edge infrastructures. In this section, we discuss the key baseline technologies, including abstraction, knowledge graphs, communication mechanisms, and others, that can provide the foundation for implementing CEI.
\subsection{Abstraction}\label{sec:baselineTech:abstractionEvolution}
In layman terms, abstraction is simply the process of hiding some complex information and allowing the system to operate with minimal necessary input from the human. For example, high level programming languages hide the complexity of interacting with the hardware and each and every assembly level executions. The primary advantage of abstraction is that it helps manage complexity. With multiple layers of abstraction, developers and system administrator can design and implement complex systems by breaking them down into manageable, interconnected, independent components/modules. Some of the benefits lies in portability, flexibility and simplicity. However, excessive or poorly designed abstraction can leverage the complexity and inefficiency of the whole system. Hence, it is essential to find the right balance that provides clarity and utility without compromising the efficiency \cite{wagner2008abstractness}. The concept of abstraction can be applied in various contexts, primarily in hardware, software, network, data, and functionality, as summarized below: 

\begin{itemize}
    \item \textit{Hardware abstraction:} Hardware abstraction is the process of hiding hardware-specific details, offering a generalized interface for the interaction of software \cite{simmann2024design}. It ensures software can function across diverse heterogeneous hardware components without the need of specification of each component provided by the corresponding vendor. Operating System (OS) and device drivers are some of the de-facto examples of Hardware Abstraction Layer (HAL). An OS enables software to smoothly interact uniformly with the underlined hardware. On the other hand a device driver bridges a specific hardware component from a specific vendor to the broader software system, ensuring standardized interactions. 
    \item \textit{Software abstraction:} The Modern software system is becoming complex day-by-day. To make is simplified and reduce its complexity for the end user, especially the developer, introducing multiple layers of abstraction, also referred to as software abstraction (SA), is the key. SA allows developers to interact with the system on a higher level, without understanding the underlying architecture \cite{lercher2024microservice,bartolini2023stream}. For instance, developers can use high-level general-purpose programming languages, including Python\footnote{\url{https://www.python.org/}}, Java\footnote{\url{https://www.java.com/en/}}, C++\footnote{\url{https://isocpp.org/}}, etc., to write code for a variety of applications, without any knowledge about the underlying architecture. Some other examples of SA can be an API, libraries like jQuery, graphics engine like OpenGL, etc \cite{lercher2024microservice}.
    \item \textit{Network abstraction:} In a similar fashion, abstraction can also be applied to the network resource management. The network abstraction (NA) can be implemented and achieved by introducing higher-level interface for network-related operations like routing, switching, multicasting, tunneling, etc \cite{silva2024path}. The application of NA can be found in a wide variety of domains,, including network protocols, wireless network, cloud network, virtual network, etc \cite{krishnan2023openstackdp}.
    \item \textit{Data abstraction:} In the era of data as next generation gold, its difficult to understand and get insight into the huge amount of unstructured data that comes from large-scale IoT infrastructure and other computing systems. When abstraction is applied to the data, we no longer deal with the raw data and the data container, e.g. stack, queue, linked ilst, etc., rather the knowledge or the information \cite{cook2009understanding, ghrab2023core,shirali2024llm}. This is where Machine Learning (ML) and AI related advanced tools come into the picture. DA is especially implemented in the modern database management system. High-level programming languages can be used to implement DA layer for any modern data-intensive application \cite{shirali2024llm}. For example, the concept of ``abstract'', ``interface'' can be used to create custom data type for real-world physical objects. 
    \item \textit{Functional abstraction:} Functional abstract is introduced to hide complexity of an algorithm which is designed to perform a specific task so that the use can simply focus on high-level business logic. A user should only pay attention to the calling convention and assume that the function will return the desired result \cite{villagrossi2023hiding}.  As an example, in the \textit{sqrt} function in Java or Python or any mathematical related library, implementation details are abstracted from the exact steps involved in finding the square root of any integer value. 
\end{itemize}

\subsubsection{Layers of abstraction in an edge device}
Figure \ref{fig:abstraction_layers_edge_device} shows different layers of abstraction in a typical edge device capable of preprocessing the input data based on either AI or non-AI-based business logic. Each layer of abstraction hide certain level of complexity making it easier to collect and interact surrounding intelligence. A brief overview of each layer is presented below: 
\begin{figure}[h]
    \centering
    \includegraphics[width=0.99\linewidth]{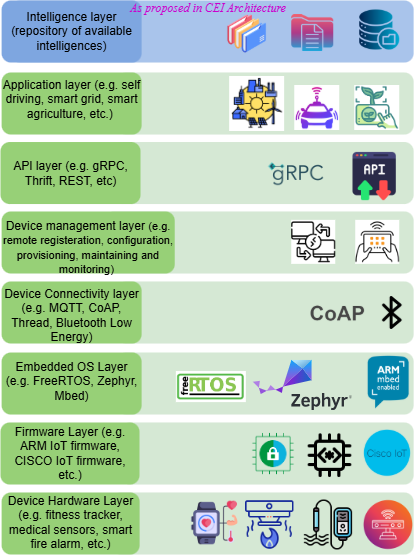}
    \caption{Different layers of abstraction in an edge device.}
    \label{fig:abstraction_layers_edge_device}
\end{figure}
\begin{itemize}
    \item \textit{Device Hardware Layer}: This is the foundational layer of any edge device, including fitness trackers, medical sensors, and smart fire alarms, consists of the physical components and hardware modules \cite{kong2022edge}. For example, it includes the processor, memory, storage, attached sensors (e.g. MPU-6050 gyroscope and accelerometer sensor, HC-SR04 ultrasonic distance sensor), and interfaces such as USB and Wi-Fi.
    \item \textit{Firmware Layer}: Positioned just above the hardware layer, the firmware layer consists of software that provides low-level control over the device's hardware. Examples include ARM IoT firmware and CISCO IoT firmware\footnote{\url{https://developer.cisco.com/docs/iotod/}}. Essentially, firmware bridges the gap between hardware and other software tools, ensuring hardware components can execute more complex tasks directed by high-level software layers.
    \item \textit{Embedded OS Layer}: This layer incorporates the OS specifically tailored for embedded systems. Such operating systems are optimized for performance and resource usage in devices with constraints on memory, processing power, or battery life. Examples include FreeRTOS\footnote{\url{https://www.freertos.org/}}, Zephyr\footnote{\url{https://www.zephyrproject.org/}}, Mbed\footnote{\url{https://os.mbed.com/}}, and Raspberry Pi OS\footnote{\url{https://www.raspberrypi.com/software/operating-systems/}}. These operating systems manage hardware resources and provide essential services for the software applications above.
    \item \textit{Device Connectivity Layer}: Serving as the communication layer, it focuses on protocols and technologies that facilitate the device's ability to connect and communicate with other devices or networks. Protocols such as MQTT\footnote{\url{https://mqtt.org/}}, CoAP\footnote{\url{https://coap.space/}}, and BLE ensure seamless data transfer and device interoperability. The MQTT protocol, for example, allows IoT devices and the cloud to communicate with one another.
    \item \textit{Device Management Layer}: This layer emphasizes the administrative aspect of edge devices. It encompasses functionalities like remote registration, configuration, provisioning, and monitoring of the device. It ensures that devices function efficiently, securely, and can be updated or diagnosed remotely.
    \item \textit{API Layer}: Standing for Application Programming Interface, this layer offers a set of routines, protocols, and tools that allow for the building and interaction of software applications. Interfaces such as gRPC\footnote{\url{https://grpc.io/}}, Apache Thrift\footnote{\url{https://thrift.apache.org/}}, and REST ensure that different software components can communicate and work together, providing a consistent and unified approach to accessing device functionalities.
    \item \textit{Application Layer}: This layer includes end-user applications that run on the edge device. The applications are designed for specific tasks like self-driving functionalities, smart grid management, smart agriculture practices, and more. They utilize services and functions from all the underlying layers to deliver specific user-centric functionalities.
    \item \textit{Intelligence Layer}: This top layer is envisioned as a core capability of the proposed CEI architecture. It maintains and manages the intelligence available on an edge device, regardless of whether that intelligence is generated through an AI model, a non-AI algorithm, a rule-based mechanism, or custom business logic. By abstracting intelligence from the underlying complex implementation and operations, this layer enables intelligence to be identified, published, discovered, shared, reused, and consumed by other edge agents.
\end{itemize}

\subsection{Knowledge Graph}
At its core, CEI aims to abstract the intelligence from 
devices, which will enable discoverability, contextual linking, 
and collaborative reasoning across edge nodes. However, 
representing intelligences (e.g., $CO_2$ level trend, temperature 
anomaly) as nodes, connecting them via semantic relationships (e.g., affects, correlatesWith, dependsOn) and capturing temporal, spatial, and contextual metadata need the use of knowledge graph as one of the baseline technology \cite{hogan2021knowledge}. 

Further, while implementing the edge inventory component, which tracks what intelligences are available (e.g., ``Average $CO_2$ in Zone B''), which edge device provides it and the corresponding metadata, one candidate solution to implement is the use of a simple key-value pair. However, this approach lacks semantic understanding (e.g., what is $i1$ about in Figure \ref{fig:dcc-vs-icc}?). The discovery mechanism would also be entirely based on the intelligence ID and not context-based (i.e., ``What's relevant to air pollution?'' is not possible to derive). With such a simple key-value pair-based approach, there is no way to relate different intelligences (e.g., ``$NoiseLevelHigh$'' $->$ ``$CO_2LikelyHigh$'').

Google recently introduced Open Knowledge Format (OKF) \footnote{\url{https://cloud.google.com/blog/products/data-analytics/how-the-open-knowledge-format-can-improve-data-sharing}}, which is an open and vendor-neutral specification introduced for representing and exchanging knowledge in a form readable by both humans and AI agents. This recent development may complement and can be used as the enabling technology for intelligence sharing among edge agents. In the context of CEI, an OKF-inspired representation could be used to describe intelligence together with its metadata, such as type, description, provider, timestamp, location, confidence, access endpoint, and others. However, OKF provides a general knowledge representation format rather than a complete intelligence-management solution; therefore, CEI-specific extensions would still be required to capture specific properties such as freshness, clustering relationships, provenance, spatial and temporal validity, lifecycle state, etc.


\subsubsection{Knowledge-data mapping}
At the edge, raw sensor data is meaningless unless it is contextualized, as discussed before. To participate in intelligence clustering, sharing, or reasoning, the device and the agent deployed on top of it must know what its data means - and map that data into a knowledge graph node (a semantic description). Without this, devices would just generate and share numbers with each other (e.g., $45$, $97$), without understanding the meaning of ``Temperature'' or ``PM2.5 pollution''. Thus at the device level, mapping of data to the corresponding knowledge graph would help in sharing a meaningful edge intelligence within the CEI system.

\subsection{Communication}
For a smooth collaboration among edge agents in a clustered edge intelligence (CEI) ecosystem, communication is the backbone. The communication protocols used in these systems must be lightweight, resilient, and low-latency to meet the demands of dynamic, resource-constrained edge environments. Several communication methods that can be employed in a CEI system to facilitate peer-to-peer intelligence sharing and overall system coordination includes:
\subsubsection{REST}
REST (Representational State Transfer) is a widely-used architectural style that enables communication between systems over HTTP \cite{dehury2026mastering}. In the context of CEI, REST can be used for device-to-cloud communication. While RESTful APIs are not optimized for real-time communication, they are effective for scenarios where simplicity and scalability are crucial, such as periodic updates from edge devices to the cloud or between devices sharing non-time-critical data \cite{dauda2024survey,sarasola2024iiot}.

\subsubsection{Pub-Sub}
The Publish-Subscribe (Pub-Sub) model is particularly useful in environments where devices need to communicate asynchronously. In CEI, edge agents can act as either publishers or subscribers. A publisher sends messages (data, intelligence) to a channel, while subscribers receive updates from the channel without knowing the identity of the publisher \cite{lazidis2022publish}. Pub-Sub systems, such as MQTT or CoAP, are ideal for CEI systems, where edge agents need to broadcast information (such as sensor readings or alerts) to multiple other devices or services without direct, one-to-one communication on a real-time basis. 
\subsubsection{5G/6G and LoRAWAN}
Advanced communication technologies like 5G and 6G are becoming critical for supporting high-speed, low-latency, and wide-coverage communications in edge computing networks. These technologies are capable of handling the vast amounts of data generated by edge devices, ensuring that the exchange of intelligence by edge agents is fast and reliable \cite{gupta20216g,huang2023semantic}. For example, in a smart city application, edge agents distributed across a large geographic area may use 5G or 6G networks to send and receive data about traffic, pollution levels, or emergency responses. Further,  LoRaWAN (Long Range Wide Area Network) ensures that intelligence can be shared among devices even in isolated areas, thanks to its design for low-power and long-range communication.


\subsection{Discoverability } \label{sec:baselineTech:Discovery} 
Discoverability is one of fundamental enablers in CEI. In contrast to traditional edge and IoT systems (i.e., where discovery mechanisms primarily focus on devices, services, or network endpoints \cite{murturi2019edge, murturi_decentralized_2022}), CEI requires mechanisms that treat intelligence itself as a first-class, discoverable, and observable (more on observability in next subsection) entity. This shift is essential because CEI operates in highly dynamic, large-scale environments where intelligence is continuously generated, updated, expires, or migrates across heterogeneous edge devices.  

Discoverability in CEI refers to the ability of edge agents and controllers to identify the intelligence that exists, where it is available, and under what context it is relevant. Unlike conventional service discovery, which relies on static identifiers such as service names or IP addresses, intelligence discovery must be semantic- and context-aware. These approaches have been shown to scale well for cloud-native services \cite{dragoni2017microservices,burns2016borg}, but they lack native support for semantic reasoning, contextual filtering, and confidence-aware selection, which are required in intelligence-centric systems. To address these limitations, CEI aligns with research in semantic IoT and context-aware discovery, where resources are described using metadata and semantic annotations rather than raw identifiers. Sheth et al. demonstrated that semantically enriched discovery significantly improves interoperability and automated reasoning in sensor-rich environments \cite{sheth2008semantic}.  Such works support the need for intelligence descriptors that include category, spatial scope, temporal validity, and dependencies on other intelligences.


Large-scale distributed systems rely on dedicated discovery mechanisms operating across different architectural layers, including service, device, and network discovery. Such mechanisms can be implemented using either a centralized or a distributed (client–server) model, depending on scalability, latency, and fault-tolerance requirements. The following subsection describes these discovery mechanisms and discusses their relevance in the context of intelligence-centric edge systems.


\subsubsection{Service Discovery}\label{sec:Discovery:serv}
\begin{figure}
    \centering
    \includegraphics[width=0.95\linewidth]{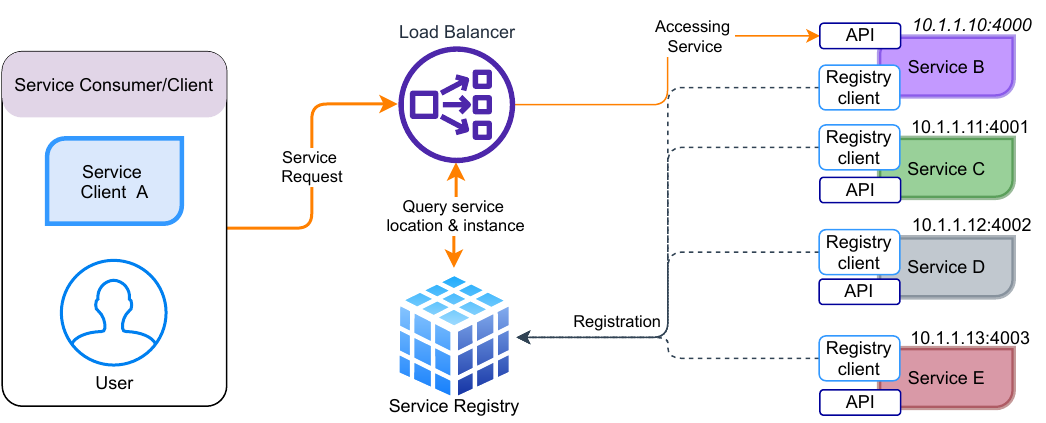}
    \caption{A general architecture of service discovery}
    \label{fig:serviceDiscovery}
\end{figure}

Service Discovery (SD), commonly used in distributed systems, is the process of automatically discovering and locating a (or a set of) service(s) based on the request made by the user or another service \cite{czerwinski_architecture_1999}, as shown in Figure \ref{fig:serviceDiscovery}. The request may comprise functional and non-functional requirements, as well as a service description. A \textit{Service Registry} (SR) is typically implemented as a database of all services, including their network locations and service instances. It is the service provider's responsibility to register their services with the service registry. The service registry can be implemented on the client or server side. Irrespective of the implementation, SR should be highly available and provide real-time information on the health and availability of all services. Several open-source and commercial service discovery solutions are available in the market. 

Some of the popular SD solutions are described below:

\begin{itemize}

 \item \textit{Docker Hub\footnote{https://hub.docker.com/}}, a server-side discovery service, is an essential component of the Docker ecosystem, providing a central location for managing and sharing Docker images. Users can create and manage their own repositories for Docker images, as well as search for and download public images created by other users. When a user searches for or pushes an updated Docker image through the Docker Command Line Interface (CLI) or REST API, the request is sent to Docker Hub's servers. Upon receiving the request, Docker Hub uses its server-side search to find the requested image or stores the newly created image in the appropriate repository. Additional features help streamline Docker container development, including automated builds, webhooks, and access control.

 \item \textit{Apache ZooKeeper\footnote{https://zookeeper.apache.org/}}, a server-side discovery service and an open-source Apache project, offers a centralized and reliable way for distributed systems to coordinate and synchronize with each other. It provides a hierarchical namespace, similar to a file system, where clients can store and access data. Clients can use ZooKeeper to discover service locations, monitor configuration changes, and synchronize data across multiple nodes in a distributed system. Coordination services are provided for client applications by a set of ZooKeeper servers. ZooKeeper's server-side infrastructure coordinates and synchronizes interactions between client applications and the distributed coordination service.
 
\item \textit{Hysterix\footnote{https://github.com/netflix/hystrix}} is an old project created to fulfill Netflix's needs. Hystrix, a client-side discovery service, is a latency- and fault-tolerance library designed to isolate access to remote services, systems, and libraries. By doing this, Hysterix protects against and controls potential latency and failures caused by external dependencies. In addition, Hysterix facilitates monitoring, alerting, and operational control in near real time. Hysterix relies primarily on manual and pre-configured settings and is not suitable for modern and real-time applications.

\item \textit{HashiCorp Consul}\footnote{https://www.consul.io/} is a service discovery, configuration, network automation, and orchestration tool. It provides a centralized registry of services and their metadata, allowing clients to discover services, securely establish communication among services, automate network routing, and control service access. When a new service is added to a network, it registers itself with the Consul agent running on each node. The agent sends the metadata of the newly added service, including IP address, port number, and health status, to the Consul server, which stores it in its registry in key-value pairs. Such data can be updated in real time without service interruption caused by a restart event. Clients can then use Consul to discover the service location using the service name. Consul can also be used for distributed locking, enabling coordination between services in a distributed system.

\item \textit{Etcd}\footnote{\url{https://www.cncf.io/projects/etcd/}}, an open-source tool, is a distributed key-value store popular for its reliability, consistency, and scalability features. This is used for implementing a configuration registry and service discovery for distributed systems, such as Kubernetes. Similar to ZooKeeper and Consul, Etcd is a server-side discovery service that provides a centralized mechanism for distributed systems to coordinate and synchronize.
 \item \textit{Traefik\footnote{https://traefik.io/}} is an open-source tool designed to provide automatic service discovery and incoming request routing ability to backend services. Request routing is based on a set of predefined rules configured manually. This can be integrated with other service registries, including the above-mentioned Consul, Etcd. In addition, this can be considered a reverse proxy and load balancer. 
 
 \item \textit{SkyDNS\footnote{https://github.com/skynetservices/skydns}} is another popular server-side discovery tool commonly used with other tools in the Kubernetes ecosystem. In the SkyDNS service discovery and management solution, containerized microservice instances are given a unique Domain Name System (DNS) name and a mapped IP address. Users send the service name, and SkyDNS returns the mapped IP address to use for communication.  
\end{itemize}

\subsubsection{Device Discovery}\label{sec:Discovery:dev}
Device Discovery (DD) is the first step toward enabling a successful Device-to-Device (D2D) communication. It is the process of finding, locating, and eventually establishing communication among nearby devices or devices connected to LAN, WAN, or the Internet \cite{hayat_device_2019}. A Device can be a computer, printer, router, switch, smartphone, or any other IoT device with a unique IP address for communication. 

Similar to the SD, DD can be of two types based on its implementation: (a) centralized and (b) distributed. In a centralized DD, a central server (e.g., a base station or access point) assists in discovering, monitoring, and managing all connected devices. The centralized entity maintains a database of all the devices, including their unique IP, MAC address, hostname,  device status, and other relevant information. The discovery client sends the discovery request to this centralized entity and acquires the required information. On the other hand, in distributed DD, each device is responsible for performing discovery by periodically sending its own information, including IP address, MAC address, and other relevant data, to other devices on the same network. Other devices respond to this information based on the requirement and update the local database if there is any update \cite{baccelli_design_2012}. 

Some of the recent use cases where DD is critical are health care services, smart home applications, industry 4.0, Industrial IoT, smart transportation, VANETs, and emergency services \cite{ccori_device_2016}. A wide range of protocols and tools are used for device discovery, including Simple Network Management Protocol (SNMP), Ping, Internet Control Message Protocol (ICMP), and network scanner tools such as Nmap \footnote{https://nmap.org/} and SolarWinds \footnote{https://www.solarwinds.com/}.
 
\subsubsection{Network Discovery}\label{sec:Discovery:network}
Unlink DD, Network Discovery (ND) focuses on knowing the entire network including the devices and all the pairs of existing connections among the devices. It is also essential to know if there is no connection between a pair of devices \cite{beerliova_network_2006}. It allows the network administrator to identify and map the topology of the network, understand the flows of data through the network, and identify potential bottlenecks or vulnerabilities \cite{ansari_5g_2018}. The topology of the network can be obtained in several ways, such as (a) by broadcasting discovery messages to neighboring devices with information about their network interfaces, IP addresses, and other relevant information, (b) by using a scanning tool, such as \textit{nmap}, to scan all connected devices and accessing device information, and (c) by analyzing the network traffic which provides the source and target device information. 

\subsection{Observability}\label{sec:baselineTech:observability}

Complementary to discoverability (as discussed in Section \ref{sec:baselineTech:Discovery}), observability is another important baseline capability for realizing CEI. Observability, in conventional distributed< systems, refers to the ability to understand and obtain the insights of the internal state and behavior of a system by continuously observing or collecting the available signals, such as logs, metrics, and traces~\cite{kosinska2023toward}.

Observability can be seen in a wide range of domains ranging from microservices~\cite{faseeha2025observability, li2022enjoy, gomes2025systematic}, to containers and from cluster of computing nodes to edge-cloud distributed computing continuum~\cite{costa2023achieving, pujol2023edge, sedlak_service_2026}. In such systems, observability is mainly used to observe the systems behavior under different circumstances and loads, monitor service health, execution latency, resource utilization, failures, and communication bottlenecks \cite{amgothu2025observability}. 

However, this conventional observability system is not sufficient for CEI, because the primary entity being managed is not only a IoT sensor, edge device, service, network endpoint, or deployed AI models and business logic, but the intelligence produced and consumed across heterogeneous edge agents in a distributed computing continuum. It must understood and implemented as intelligence-level observability. This must have the capabilities to continuously observe and collect signals related to the state, quality, reliability, and usability of available intelligence. Since intelligence in CEI is treated as a first-class entity, it is not enough to know whether an edge device is active or whether a communication link is available. The observability system must provides insights on whether a particular intelligence is fresh, valid, reliable, semantically consistent, and suitable for the current context. For example, in a fire-detection intelligence cluster, the intelligence related to recent $CO_2$ increase, temperature rise, humidity change, and air-quality variation must be observable in terms of their generation time, update frequency, confidence score, and current availability.

It is to be noted that discoverability (as discussed before) and observability are different and complementary to each other. Discoverability answers the question: \emph{``Which intelligence is available and where can it be accessed?''}. On the other hand, observability answers a different question: \emph{``Can this intelligence be trusted and used at this moment?''}. An edge agent may discover multiple sources of intelligence that provide similar information, such as traffic congestion level in a specific road segment. However, the selection of the most suitable source depends on observable properties such as freshness, confidence, latency, and provenance. Similar conclusions have been reached in knowledge-centric networking and fog computing research, where observable and semantically described knowledge artifacts outperform static, data-centric approaches in dynamic environments \cite{bonomi2012fog,zhou2019edge,pujol2023edge}.

There are several tools/implementation available in the research and industrial practice that can be used as the baseline. This includes, OpenTelemetry\footnote{https://opentelemetry.io/}, MLflow\footnote{https://mlflow.org/}, Dynatrace\footnote{https://www.dynatrace.com/monitoring/platform/observability-solution}, AgentTrace\footnote{https://github.com/tensorstax/agenttrace}. However designing and implementing observability within CEI architecture is still an open-ended research question and needs further investigation. Some of the key research challenges include deciding between fully centralized and distributed observability mechanisms, design of lightweight, hierarchical, and adaptive observability mechanism, and design of standard observable attributes for intelligence.

\subsection{Clustering mechanisms}
The heart of CEI is the mechanism on how to efficiently cluster the intelligence components that are available at the edge through large number of edge devices. To the best of our knowledge, no specific clustering algorithm is developed to perform the same for inteligence. However, we believe that the existing clustering mechanisms, such as K-means, Affinity Propagation (AP), Mean-shift, Spectral clustering, Ward hierarchical clustering, Agglomerative clustering, DBSCAN \cite{ester1996density}, OPTICS, Gaussian mixtures, BIRCH \cite{zhang1996birch}, etc. can be used as the baseline for design and development of clustering mechanisms only for intelligence. Further, it is believed that not a single clustering approach can address the problem where, the clustering can be done based on different context, type and other parameters. Centroid-based methods, such as mini-batch K-means and distributed K-means variants (e.g., k-means), are commonly employed when the approximate number of clusters is known. However, for scenarios where intelligences may exhibit irregular relationships, noise, or varying cluster densities, hierarchical density-based methods, such as scalable or approximate variants of HDBSCAN can be adopted over k-nearest-neighbor graphs, as they can discover clusters of arbitrary shape while filtering out weak or noisy intelligences. In highly dynamic environments where intelligences continuously appear, evolve, or expire, streaming clustering frameworks based on micro-cluster and macro-cluster abstractions (e.g., CluStream-, DenStream-, or ClusTree-style approaches) can be adopted. Since the research is in its very early stage, how the existing clustering mechanisms can be used as baseline is still needs to be investigated. 


\subsection{Other baseline technologies} 

Before intelligence can be clustered, it must be developed, deployed, managed in an effective way. This management spans the entire intelligence lifecycle - from development and packaging to deployment, execution, update, and uninstall. All the operations require supporting technologies that are already well matured, including IaC, containerization, orchestration, blockchain for security \cite{dehury9748997} and several others. Infrastructure-as-Code (IaC) can be used to declare the specification of intelligence, including  the underlined dependencies of hosting environments, such as edge and cloud resources, as a code. Containerization technologies can be used to package intelligence, along with the corresponding AI models, algorithms, or rule-based logic and their runtime dependencies. On top of this, orchestration frameworks can be used to coordinate the deployment, scaling, migration, and termination of intelligence instances across edge-cloud computing continuum. 

\section{Research Dimensions}\label{sec:research_dir}

\subsection{Intelligence marketplace (IM)}
\emph{Intelligence Marketplace}, a.k.a Edge Intelligence Marketplace (EIM), can be considered as a  platform or a ecosystem where intelligence (derived using AI-based models, algorithms, or even if-else rules) can be exchanged, bought or sold between the Edge Intelligence Producer (EIP) and the buyer. On the contrary, the similar term \emph{marketplace intelligence} refers to the process of collection and making use of the data and knowledge related to different marketplaces, e.g. data marketplace and e-commerce marketplace, The discussion in this paper revolves around the concept of EIM (and not \emph{marketplace intelligence}), different potential stakeholders and their interests and interactions. 
For example, a city government for traffic management may buy intelligence (based on AI models or algorithm) that predict congestion based on vehicle counter and optimize traffic light timing on a real-time. Such intelligence can be remotely deployed on the edge device inside traffic light by the intelligence provider (IP) or seller. Another example in the context of smart agriculture is that a farm operator might buy and deploy rule-based or AI-based intelligence for optimal prediction of watering times based on weather data and soil moisture sensors and control basic irrigation. The farm operator may remotely deploy this intelligence on an edge device installed near the farmland.

\begin{figure}[h]
    \centering
    \includegraphics[width=.5\textwidth]{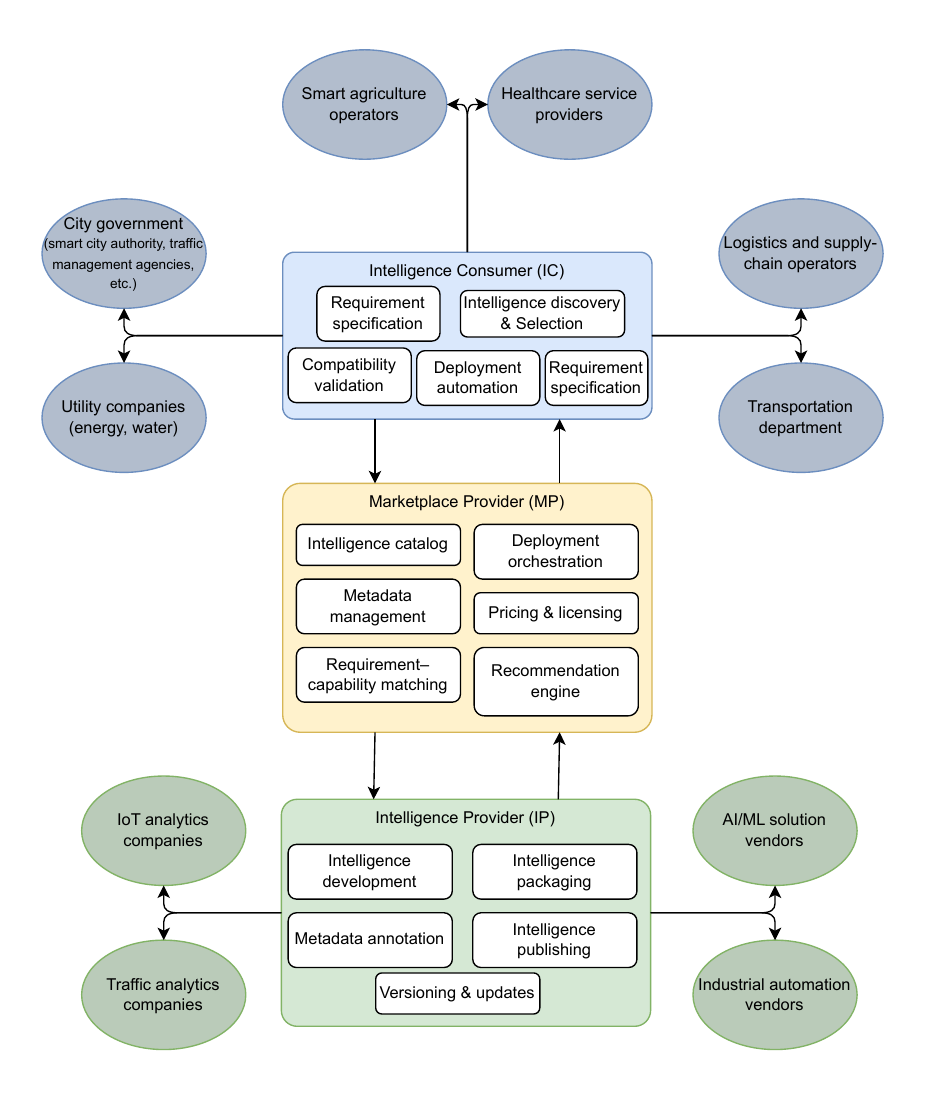}
    \caption{Interaction of stakeholders within Intelligence Marketplace }
    \label{fig:IntelligenceMarketplace}
\end{figure}

Potential stakeholders of Edge Intelligence Marketplace, as shown in Figure \ref{fig:IntelligenceMarketplace}: 
\begin{itemize}
    \item \textit{Intelligence Provider (IP)}: These can be developers, companies, or individuals who create AI models, new algorithms, or decision rules. They publish these on the marketplace. For examples, IoT and sensor data analytics companies (e.g. PTC (ThingWorx), Siemens MindSphere, or Cisco IoT) can provide/sell intelligence such as predictive maintenance of sensors, edge devices, anomaly detection in edge device, etc. Traffic and transportation analytics firms may provide/sell AI-models based intelligence that give the current traffic congestion and insights. 
    
    \item \textit{Intelligence Consumer}: Intelligence Consumers (IC) typically the business entities or individuals or governments have existing infrastructure and systems in place (e.g., sensors, edge devices and infrastructure). Such consumers may lack deep AI expertise or resources to build models from scratch and hence rely on pre-built intelligence from the marketplace to solve their specific problems. They need specific domain-specific intelligence that is aligned with their operational goals or industry challenges. IC browses the marketplace and select the intelligence (whether it is based on a complex AI model, a simpler algorithm, or an if-else rule set) that suits their needs and compatible with their existing edge infrastructure.  
    
    \item \textit{Marketplace provider}: 
    The Marketplace Provider (MP) can be seen as a trusted platform where intelligence consumer and provider gathered. Intelligence provider (IP) lists the intelligence with a set of metadata and and pricing/licensing terms. In addition, the IP should also be responsible for updating and maintaining intelligences. On the other hand, intelligence consumer would specify his/her requirements, search and select intelligences, acquire usage rights, deploy them onto its edge resources, and utilize their outputs for application-level decision-making. The MP could be seen as a combination of e-commerce platform (like Amazon) and model hub (like Hugging Face Hub) with the additional capabilities of matching consumer's requirement and providers capabilities, recommendation system for both IP and IC and an cost-analytics system. 
\end{itemize}
    

\subsection{Intelligence discoverability \& observability}
In a system with minimal human intervention, an edge agent must be capable of discovering other relevant intelligences in order to cross-verify and validate its own intelligence before publishing it. Furthermore, when new intelligence becomes available within the edge infrastructure system, an edge agent should be able to utilize it to improve the reliability and accuracy of its own intelligence. A key research challenge, however, lies in how to design, develop, and implement an intelligence discovery mechanism in resource-constrained environments. Existing discovery mechanisms-whether service discovery, device discovery, or network discovery-cannot be directly adopted for intelligence discovery. Another major challenge is the observability of intelligence, such as monitoring its availability and freshness. Designing a fully distributed observability mechanism, where each edge agent monitors other intelligences, would be both network- and computation-intensive. Conversely, a fully centralized observability system introduces inherent and obvious challenges, including single points of failure, scalability bottlenecks, and increased latency. 

\subsection{Intelligence lifecycle automation}

Similar to lifecycle automation of software systems in production environment, intelligence lifecycle involves 
development (using AI models, algorithms, or rule-based logic), packaging, deployment to edge devices or controllers, configuration, execution, monitoring, update or adaptation, and finally uninstalling the intelligence. Automation can be applied on multiple levels of this lifecycle, from basic deployment and start/stop operations to advanced capabilities including automated scaling, migration, versioning, rollback, and context-driven reconfiguration based on environmental changes~\cite{sedlak_service_2026}. While implementing such automation the potential research challenges that we may encounter are how to maintain consistency, how to handle inter-dependencies among multiple intelligences, and manage heterogeneous device capabilities. The implementation of automation become more cumbersome in coordinating lifecycle actions across distributed edge agents while maintaining low latency, scalability, and fault tolerance.

The existing orchestrators are matured enough to carry out the lifecycle operations. However, such solutions are not suitable for performing the lifecycle operations on the resource devices at the edge of the network. One approach to address this research challenge is to ensure the aumation while the orchestrator being deployed on the cloud environment. However, this will not be an distributed solution and hence issues such as network stability may become an obstacle in ensuring lifecycle automation. The other approach is to design the orchestrator small enough so that it can be installed on the resource constraint edge devices and responsible to taking care of each edge device separately.


\subsection{Clustering mechanism}

This is one of the primary focus area, as an efficient clustering mechanism would determine how individual intelligences are grouped, managed and leveraged for collaborative decision-making. The clustering may follow a manual process, a policy-driven approach defined by administrators, an algorithmic approach based on classical clustering mechanisms, ML methods, or Large Language Model (LLM)-assisted semantic reasoning. The open-ended research questions are where clustering logic resides and who decides the clusters? Clusters may be formed locally by the edge agents, coordinated by the edge controllers or gateways or orchestrated at the cloud level depending on the latency and scalability. Another research aspect is the nature of the cluster itself: in CEI, clusters do not necessarily have a physical manifestation and are primarily logical/virtual constructs that represent relationships among intelligences rather than underlined co-located devices. Such logical clusters may be instantiated, maintained, or visualized at different layers of the edge-cloud computing continuum. Considering above questions, the research challenges may include designing adaptive clustering mechanisms that can operate across computing layers and ensures that the clusters are meaningful, consistent and dynamic. 
    
    

\subsection{Use of language models}
Use of language model could be one potential solution to identify and establish relationships among multiple intelligences (e.g., which intelligences participate in remote fire detection or traffic congestion estimation) instead of manually curating clusters. LLM can fill the gap where CEI needs context-aware, multi-constraint clustering and discovery over several artifacts, including AI models, algorithms, rules. LLMs can help in translating natural-language intents (e.g., ``form a fire-detection cluster for forest zone $X$ with freshness $<$ 2 min and confidence $>$ 0.8'') into structured queries over inventories/knowledge graphs using specific language. Modern language models can also help in figuring out how intelligence from edge agents are related (like what affects what (\textit{correlatesWith}), what relies on what (\textit{dependsOn}), and what confirms what (validates)) to suggest candidate clusters and justifications.

Language models can be used to complement the formation of clusters. This means LLM can help us find the relationship among multiple intelligence. For example, what are the intelligences that can take participate in detecting the fire in a remote location. Another example is ``what are the intelligence that can be considered to estimate the traffic congestion?'' 

\subsection{Interoperability and Standardization}
As discussed before, there could be multiple IPs and ICs. In such a situation, intelligences may be generated by different vendors, implemented using different AI models, algorithms, or rule-based logic, and executed on edge devices with varying hardware capabilities, OS, and communication protocols. Intelligence management will become fragmented and difficult to scale across computing continuum if there is no common standard for intelligence representation, description of metadata, communication interface and protocols, and lifecycle operations. For example, the IP publishing the traffic congestion related intelligence may expect the vehicle count data in a JSON format for MQTT. Similarly, the IP publishing pollution monitoring-related intelligence may expect the $CO_2$ data in a different schema over REST. Without standardized data formats, interface, and metadata description, these two intelligences cannot be combined or clustered. 


\section{Use cases}\label{sec:usecases}

\subsection{Industry 4.0}
In industry 4.0, such CEI can be applied on a wide range of activities, including predictive maintenance cluster for CNC machines, conveyor belt fault validation cluster, robotic arm collision-risk cluster, industrial fire and gas leak detection cluster, compressor health check, vibration-temperature-pressure anomaly cluster, industrial wastewater discharge compliance cluster, and many more \cite{CARVALHO2019276}.

For example, one edge device attached to a gas sensor may publish intelligence like ``abnormal rise of methane in Zone B''. Another edge device with a thermal camera may publish ``heat anomaly near valve section''. A third edge device may publish ``real-time increase of smoke'', while a fourth edge device may publish ``unstable ventilation detected''. These separate intelligences can be clustered semantically. If gas anomaly, heat anomaly, and smoke-related intelligence appear together within the same time window and location, the cluster can raise a high-confidence ``probable fire/gas leak incident in Zone B'' alert.  Similarly, if vibration anomaly combines with unusual temperature rise and pressure deviation from the same machine section, the cluster may infer ``high likelihood of mechanical degradation'', ``possible bearing failure'', ``seal leakage'', or ``pump cavitation risk'', depending on the context and rules used.

\subsection{Smart City \& Traffic Management System}
Similar to Industry 4.0, the potential of CEI can be seen on a wide range of smart city and traffic management related activities, including pedestrian crowding risk cluster, flooded road detection cluster, road accident confirmation cluster, emergency vehicle priority cluster, illegal parking validation cluster, traffic congestion estimation cluster, low-visibility hazard cluster, intersection collision-risk cluster, and many more \cite{fi17030118}. For instance, an edge device with roadside camera may publish intelligence like ``high pedestrian density near crossing A''. Another edge device at a smart traffic signal may publish ``extended pedestrian waiting time''. A third edge device near a bus stop may publish ``unusual crowd buildup'', while a fourth edge device may publish ``sharp increase in nearby device presence''. Clustering the above intelligence, a high-confidence ``probable pedestrian crowding risk near crossing A'' event can be derived reliably. Similarly, if the same cluster also observes repeated road crossing attempts and reduced walking space, it may infer ``high likelihood of unsafe pedestrian spillover''.

\subsection{Smart Agricultural}
With recent advancements, drone technology is becoming more affordable, reliable, and efficient. Drones equipped with multiple sensors, along with complementary technologies such as satellite image analysis, are increasingly being used in smart agriculture \cite{TARIQ2025107342}. In the context of CEI, one edge device connected to a wind sensor may publish intelligence such as ``strong gust activity detected in Plot A''. Another edge device attached to a weather station may publish ``rapid drop in atmospheric pressure''. A third edge device mounted on a crop-monitoring camera may publish ``visible crop bending beyond normal range'', while a fourth edge device may publish ``soil surface dryness reducing root grip''. These separate intelligences can be clustered semantically. For instance, if strong gust activity, pressure instability, and abnormal crop bending appear together within the same time window and location, the cluster can generate a high-confidence intelligence alert such as ``probable wind-damage risk in Plot A''.

CEI, together with drone technology, can support a wide range of agricultural activities, including waterlogging detection, crop disease early warning, rainfall–soil moisture inconsistency detection and prediction, and many more.

\subsection{Other use cases}
Beyond the above use cases, CEI can also be applied in healthcare monitoring systems, home automation, industrial safety, transportation \cite{zhang2022information}, public infrastructure monitoring, environmental protection, collision avoidance in UAV system \cite{KHARGHARIA2026132020, zhang2026learning} and many other domains. The key idea remains the same across all these use cases: instead of relying on a single edge device or a single type of intelligence, multiple related intelligences generated at the edge can be semantically clustered to produce a more reliable and context-aware outcome.

\section{Literature Survey}\label{sec:rel}

This section reviews the academic landscape around EI and CEI across three core dimensions: first, we examine the shifting taxonomies of edge intelligence, highlighting the transition to intelligence as independent assets. Second, we survey the semantic technologies required to represent and codify knowledge across heterogeneous nodes in a system. Third, we analyze the paradigm shift from hardware-bound device networks to insight-driven clusters. Table~\ref{tab:literature-overview} summarizes the literature considered.

\begin{table*}[t]
\footnotesize
    \centering
    \begin{tabular}{ccccc}
    \hline
       Ref. & Year & Focus Area  & Article Type & Topic \\
       \hline \hline
       Zhou et al.~\cite{zhou2019edge} & 2019 & Paradigms and Taxonomy & Review & Technologies and frameworks enabling EI\\
       Deng et al.~\cite{deng_edge_2020_new} & 2020 & Paradigms and Taxonomy & Vision & Overview of EI paradigms\\
       Xu et al.~\cite{xu_edge_2020} & 2020 & Paradigms and Taxonomy & Survey & Fundamental components of EI\\
       Chen et al.~\cite{chen_deep_2019} & 2019 & Paradigms and Taxonomy & Review & Implications of deep learning for EI\\
       Gill et al.~\cite{gill2025edge} & 2025 & Paradigms and Taxonomy & Survey & Collaborative EI architectures \\
       Stadnicka et al.~\cite{stadnicka_industrial_2022} & 2022 & Paradigms and Taxonomy & Review & Industrial applications of EI \\
       \hline
       Zhang et al.~\cite{zhang2022information} & 2022 & Semantics and Abstractions & Survey & Information fusion as an enabler for EI \\
       Ghrab et al.~\cite{ghrab2023core} & 2023 & Semantics and Abstractions & Review & Ontologies for IoT and Edge computing \\
       Haller et al.~\cite{haller_modular_2019} & 2019 & Semantics and Abstractions & Review & Ontologies for sensing and processing \\
       Li et al.~\cite{li_review_2019} & 2019 & Semantics and Abstractions & Review & Ontologies for Internet of Things \\
       \hline
       Dehury et al.~\cite{dehury2022ccei} & 2022 & Intelligence Clustering & Research & Emergence of CEI based on topologies \\
       Achkouty et al.~\cite{achkouty2024rdsc} & 2024 & Intelligence Clustering & Research & Intelligence clusters in IoT networks \\
       Takale et al.~\cite{takele2025resource} & 2025 & Intelligence Clustering & Research & Usage of CEI for improved FL\\
       Zhang et al.~\cite{zhang_toward_2025} & 2025 & Intelligence Clustering & Survey & Usage of agentic AI as enabler for EI \\
       Mughal et al.~\cite{mughal2024adaptive} & 2024 & Intelligence Clustering & Research & Usage of CEI for improved FL \\ \hline \hline
    \end{tabular}
    \caption{Overview of literature revolving around EI and its applications, categorized into three focus areas.}
    \label{tab:literature-overview}
\end{table*}

\subsection{Paradigms and Taxonomy of Edge Intelligence}

Existing surveys and taxonomies in EI have targeted mainly the first two paradigms discussed in this paper: EI-1F and EI-2F. Closely aligning to these two paradigms, Deng et al.~\cite{deng_edge_2020_new} distinguish between ``intelligence-enabled edge computing''  and ``artificial intelligence on the edge''. Xu et al.~\cite{xu_edge_2020} take an additional step in their survey, where they identify four cross-cutting applications of EI: edge caching, edge training, edge inference, and edge offloading. Stadnicka et al.~\cite{stadnicka_industrial_2022} conducted a large-scale industrial review where they investigated the potentials of EI for various domains. Across these potential applications, Chen et al.~\cite{chen_deep_2019} review the importance of deep learning techniques as an enabler for adaptive and robust decision-making; Zhou et al.~\cite{zhou2019edge} strike a similar chord, discussing the importance of deep learning for edge environments to split and distribute intelligence within an architecture. Lastly, Gill et al.~\cite{gill2025edge} widen the scope and describe the architectures to facilitate the interaction and collaborating of distributed software components.

Despite this compact body of work, few ideas revolve around the third paradigm discussed in this paper: EI-3F---the intelligence available at the edge, where the abstract, derived insights themselves---rather than raw data or hardware models---are treated as independent, first-class manageable entities. This work addresses this gap.

\subsection{Semantic Representation and Abstraction at the Edge}

To codify, comprehend, and exchange knowledge across heterogeneous edge environments, current research heavily leverages semantic web technologies, ontologies, and knowledge graphs. Early foundational efforts standardized the physical layer: Haller et al.~\cite{haller_modular_2019} introduced the modular \textit{SOSA}/\textit{SSN} ontologies for sensors, actuators, and observations, while Li et al.~\cite{li_review_2019} evaluated Web of Things (WoT) ontologies to drive cross-layer data interoperability and federation. As these standards matured, modeling shifted toward edge infrastructure; for instance, Ghrab et al.~\cite{ghrab2023core} proposed EdgeOnto to automate the IoT service lifecycle by mapping node capabilities and Quality of Service (QoS) constraints. Lastly, to elevate raw data into actionable EI, Zhang et al.~\cite{zhang_information_2022} highlighted context-aware and real-time information fusion as core mechanisms . 

While standardized ontologies effectively describe data sources and infrastructure, they fail to manage abstract insights as independent assets. To federate intelligence effectively, edge knowledge graphs must model these insights as distinct nodes connected by explicit semantic relationships (e.g., affects, correlatesWith) rather than flat, semantics-free key-value pairs. This shows a gap for ontologies that explicitly encapsulate metadata native to abstract insights, including provenance, confidence levels, geographic scope, temporal freshness, clustering relationships, and operational lifecycle states. 

\subsection{From Device-Centric to Clustering-Centric}

The coordination of distributed edge nodes is shifting from physical topologies to intelligence-based grouping. Traditional DCC is grouping edge nodes or IoT devices rigidly by geographic proximity, communication distance, or sensing coverage~\cite{dehury2022ccei}. For instance, Achkouty et al.~\cite{achkouty2024rdsc} and Takale et al.~\cite{takele2025resource} propose spatial clustering to manage device coverage, energy, and storage limits. While these methods optimize data aggregation and network lifetime, they treat nodes strictly as physical containers bounded by hardware and network constraints. Even distributed architectures, like FL, commonly rely on centralized cloud coordinators; this shows a need for multi-edge clustering and direct node communication~\cite{mughal2024adaptive}.
For instance, Dehury et al.~\cite{dehury2022ccei} form logical clusters based on the specific intelligence devices possess rather than their physical parameters, enabling decentralized peer collaboration with minimal energy expenditure. Furthermore, Zhang et al.~\cite{zhang_toward_2025} show how agentic AI enables edge systems to run continuous perception–reasoning–action loops, allowing nodes to cluster autonomously based on semantic relevance. By treating intelligence as a manageable asset, ICC bypasses hardware vendor lock-in and fosters decentralized collaboration across dynamic edge continuums.

In summary, we can report that DCC emerges as a powerful enabler for CEI, allowing distributed entities to act as autonomous and self-organized agents~\cite{zhang_toward_2025}, or bypass traditional centralized architectures~\cite{mughal2024adaptive}. Thus, we conclude that CEI is an emerging field with high potential, with most available works focusing on original research, rather than established surveys and reviews (Table~\ref{tab:literature-overview}).





\section{Conclusions and Future Works}
\label{sec:conc}

In this paper, we introduced Clustered Edge Intelligence (CEI) as a vision that redefine Edge Intelligence beyond the conventional use of AI for edge resource management and the deployment of lightweight AI models on edge devices. The existing edge systems provide limited support for managing derived intelligence as an independent entity that can be discovered, shared, reused, and combined across heterogeneous edge devices. The proposed CEI framework filled this gap by first interpreting edge intelligence as the intelligence available at the edge and treating such intelligence as a first-class entity. This intelligence may be derived using AI models, conventional algorithms, rule-based mechanisms, or custom business logic.

The proposed CEI framework shifts the focus from device-centric clustering to intelligence-centric clustering. Such clusters allow edge agents to combine and cross-validate intelligence generated by heterogeneous devices without necessarily sharing raw data. It is important to note that CEI is presented as a vision and reference architecture rather than as an implemented and empirically evaluated system. To support the realization of CEI, we discussed a set of baseline technologies, including abstraction, knowledge graphs, communication mechanisms, intelligence discovery and observability, clustering techniques, life cycle automation, containerization, orchestration, and intelligence marketplaces. Furthermore, we investigated several potential research directions in this emerging domain, such as intelligence-centric clustering, distributed intelligence inventories, semantic discoverability, intelligence observability, life cycle management, interoperability, standardization, and the use of LLMs for identifying relationships among heterogeneous intelligence.

Despite the wide range of potential research directions, we are currently focusing on developing a CEI prototype and defining a standard representation for intelligence, including its identity, provenance, confidence, freshness, and communication requirements. Our future work also includes designing lightweight and adaptive mechanisms for intelligence discovery, observability, and lifecycle management across the distributed computing continuum.

\subsection*{Acknowledgement}
This work is partially funded by IISER Berhampur,  institutional seed grant SG/CKD/050225. This work is also partially supported by the EU's TALENTS (101299722) project. 
\bibliographystyle{elsarticle-num}
\bibliography{references,Boris,ilir, Chinmaya}

\end{document}